%% file: ClonedPerson.tex
\documentclass[10pt,twocolumn,letterpaper]{article}

\usepackage{cvpr}              % To produce the CAMERA-READY version
\usepackage{graphicx}
\usepackage{amsmath}
\usepackage{amssymb}
\usepackage{booktabs}
\usepackage{multirow}
\usepackage{ulem}
\usepackage{tablefootnote}
\usepackage{caption}
\usepackage{import}
\usepackage[table]{xcolor}
\usepackage[labelformat=simple]{subcaption}
\usepackage{tabularx}
\usepackage{makecell}
\usepackage{docmute}
\usepackage{chapterbib}
\usepackage{subcaption}

\usepackage[pagebackref,breaklinks,colorlinks]{hyperref}

\usepackage[capitalize]{cleveref}
\crefname{section}{Sec.}{Secs.}
\Crefname{section}{Section}{Sections}
\Crefname{table}{Table}{Tables}
\crefname{table}{Tab.}{Tabs.}

\def\cvprPaperID{628} % *** Enter the CVPR Paper ID here
\def\confName{CVPR}
\def\confYear{2022}

\begin{document}

\title{Cloning Outfits from Real-World Images to 3D Characters\\for Generalizable Person Re-Identification}

\author{Yanan Wang$^1$\\
{\tt\small yanan.wang.cs@gmail.com}
\and
Xuezhi Liang$^{2}$\\
{\tt\small xz.liang.cs@gmail.com}
\and
Shengcai Liao$^{1}$ \thanks{Shengcai Liao is the Corresponding Author.}\\ 
{\tt\small scliao@ieee.org}
\and
Inception Institute of Artificial Intelligence (IIAI)$^{1}$\\
Mohamed bin Zayed University of Artificial Intelligence$^{2}$\\
Masdar City, Abu Dhabi, UAE\\
}
\maketitle

%%%%%%%%% ABSTRACT
\begin{abstract}
    Recently, large-scale synthetic datasets are shown to be very useful for generalizable person re-identification. However, synthesized persons in existing datasets are mostly cartoon-like and in random dress collocation, which limits their performance. To address this, in this work, an automatic approach is proposed to directly clone the whole outfits from real-world person images to virtual 3D characters, such that any virtual person thus created will appear very similar to its real-world counterpart. Specifically, based on UV texture mapping, two cloning methods are designed, namely registered clothes mapping and homogeneous cloth expansion. Given clothes keypoints detected on person images and labeled on regular UV maps with clear clothes structures, registered mapping applies perspective homography to warp real-world clothes to the counterparts on the UV map. As for invisible clothes parts and irregular UV maps, homogeneous expansion segments a homogeneous area on clothes as a realistic cloth pattern or cell, and expand the cell to fill the UV map. Furthermore, a similarity-diversity expansion strategy is proposed, by clustering person images, sampling images per cluster, and cloning outfits for 3D character generation. This way, virtual persons can be scaled up densely in visual similarity to challenge model learning, and diversely in population to enrich sample distribution. Finally, by rendering the cloned characters in Unity3D scenes, a more realistic virtual dataset called ClonedPerson is created, with 5,621 identities and 887,766 images. Experimental results show that the model trained on ClonedPerson has a better generalization performance, superior to that trained on other popular real-world and synthetic person re-identification datasets. The ClonedPerson project is available at \url{https://github.com/Yanan-Wang-cs/ClonedPerson}.
\end{abstract}

%%%%%%%%% BODY TEXT
\section{Introduction}
\label{sec:intro}

The generalization of person re-identification has gained increasing attention in recent years. One way to improve generalization is to develop large-scale and diverse training datasets. However, collecting person images from surveillance videos is privacy sensitive, and the further data annotation is expensive. Therefore, recently, synthetic person re-identification datasets have been actively developed due to their advantages of no privacy concern and no annotation cost \cite{barbosa2018looking, bak2018domain, sun2019dissecting}. For example, RandPerson \cite{wang2020surpassing} automatically creates large-scale random 3D characters with 8,000 identities, rendered from simulation of surveillance environments in Unity3D \cite{unity}. It is also proved in \cite{wang2020surpassing} that large-scale synthetic datasets are very useful to improve generalization. Similar findings are also observed in the following work UnrealPerson \cite{zhang2021unrealperson}. However, synthesized persons in existing datasets are quite different from realistic persons, because synthesized persons are mostly cartoon-like and dress in random collocation. This clear domain gap limits the performance of models trained on such synthetic datasets.

On the other hand, some researchers proposed to generate 3D human body models from real-world person images \cite{lazova2019360,wang2019re,zhao2020human}, targeting at high-fidelity reconstruction. These methods try to generate 3D body shapes and the associated textures simultaneously, through deep neural networks. They can help reduce the gap between synthetic and realistic person images to some extent due to the input of real-world clothes textures. However, current methods are still not satisfactory as the results are usually blur, and there are many artifacts, e.g. in back views (see \cref{fig:com_hpbtt}).

\begin{figure*}
  \centering
  \includegraphics[width=0.8\linewidth]{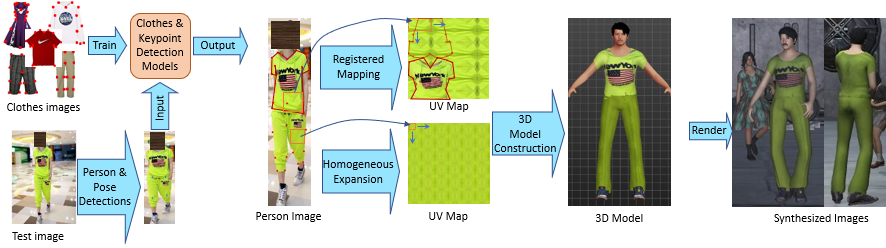}
  \caption{The proposed ClonedPerson pipeline, which automatically creates similarly dressed 3D characters from person images.}
  \label{fig:pipeline_img}
\end{figure*}

Considering the above, in this work, an automatic approach is proposed to directly clone the whole outfits from real-world person images to virtual 3D characters. By doing so, we would like to achieve two goals\footnote{However, high-fidelity 3D reconstruction of person bodies, for example, heads and 3D body shapes, is not our target. On the other hand, high-fidelity reconstruction of identifiable biometric signatures, e.g. faces, may also raise privacy concerns.}: (1) the directly cloned clothes textures are clear and sharp in looking; and (2) by cloning the whole outfit, the virtual person thus created will appear very similar to its real-world counterpart, in similar clothes and dress collocation. Specifically, inspired from the UV texture mapping \cite{chang2006modeling} method developed in RandPerson \cite{wang2020surpassing}, in this work, two cloning methods for UV maps are designed, namely registered clothes mapping and homogeneous cloth expansion. Registered mapping targets at regular UV maps where clothes appear in regular shapes and structures. Based on clothes keypoints detected on real-world person images and labeled on UV maps, registered mapping applies perspective homography \cite{szeliski2010computer} to warp real-world clothes to the counterparts on the UV map. Homogeneous expansion is for invisible clothes parts and irregular UV maps. An optimization algorithm is proposed to find a large homogeneous area on clothes, use it as a realistic cloth pattern or cell, and expand the cell to fill the UV map. \cref{fig:pipeline_img} shows the pipeline of the proposed method.

Furthermore, a general principle is established to scale up virtual 3D character creation, that is, it should expand both densely in 
% visual 
similarity and diversely in population. The former one is to challenge discriminative model learning by providing 
% visually 
similar persons, while the latter 
% one 
is to enrich the diversity in sample space. A similarity-diversity expansion strategy is thus proposed. Thanks to the proposed clothes cloning method, this can conveniently be achieved by clustering real-world person images and a controlled sampling of the clustered images for 3D character generation. 

Eventually, the generated 3D characters are imported into Unity3D virtual environments to render a more realistic virtual dataset, called ClonedPerson, with 763,953 images from 4,826 characters for training, and 123,813 images of 795 characters for testing. Experimental results show that the similarity-diversity expansion strategy is effective, and the model trained on the ClonedPerson dataset has a better generalization performance, surpassing the models trained on various real-world and synthetic datasets. 

In summary, our main contributions are: (1) We propose an automatic pipeline to clone outfits from real-world person images to virtual 3D characters, such that they look very similar to their real-world counterparts, with clear clothing textures; (2) two cloning methods, registered clothes mapping and homogeneous cloth expansion, are designed to fulfill this task; (3) a similarity-diversity expansion strategy is proposed, based on clustering of person images and controlled sampling, to scale up 3D character creation densely in similarity and diversely in population; and (4) a large-scale synthetic dataset called ClonedPerson, with 887,766 images of 5,621 characters, is created, which results in a better generalization performance than other popular real-world and virtual person re-identification datasets. All used and designed methods are listed in Table A of the Appendix.

%------------------------------------------------------------------------

\section{RELATED WORK}
\label{sec:related_work}

\begin{table}
 \begin{center}
  \begin{tabular}{@{}c@{}|c|@{}c@{}|@{}c@{}|c|@{}c@{}|@{}c@{}}
  \hline
    Dataset& \#ID & \#Cam & \#BBox & Sur & Real & Outfit \\
   \hline\hline
   SOMAset \cite{barbosa2018looking}   & 50  & 250 &100,000 &No&No&No\\
    SyRI \cite{bak2018domain}    & 100& 280&56,000&No&No&No\\
   PersonX \cite{sun2019dissecting}   & 1,266 & 6& 273,456&No&No&No\\
   RandPerson\tablefootnote{Suggested subset from \cite{wang2020surpassing}: 8,000 characters with 132,145 images.} \cite{wang2020surpassing} & 8,000& 19  &1,801,816& Yes&No&No\\
   UnrealPerson\tablefootnote{Suggested subset from \cite{zhang2021unrealperson}: 3,000 characters with 120,000 images.} \cite{zhang2021unrealperson} & 6,799 & 34 &1,256,381 & Yes&Yes&No\\
   ClonedPerson & 5,621& 24 & 887,766  &Yes&Yes&Yes\\
   \hline
  \end{tabular}
    \caption{Statistics of some synthetic person re-identification datasets. ``Sur": surveillance simulation. ``Real": realistic clothes textures. ``Outfit": cloning full-body outfits from person images.}
    \label{tab:dataset_compare}
 \end{center}

\end{table}

Collecting and manually labeling real-world person re-identification datasets are expensive and privacy-sensitive. In contrast, the use of synthetic data can reduce the cost of manual labeling, and synthetic datasets do not have privacy issues. For synthetic datasets, SyRI \cite{bak2018domain} and PersonX \cite{sun2019dissecting} used limited hand-made characters to generate data. In contrast, RandPerson \cite{wang2020surpassing} proposed a clever way to generate new-looking clothes models by replacing UV maps of existing 3D clothes models with neutral images or random color and texture patterns, and designed an automatic pipeline in MakeHuman \cite{makehuman} to scale up character generation. Besides, similar to real-world environments, \cite{wang2020surpassing} simulated camera networks in Unity3D to render and record moving person videos. Moreover, \cite{wang2020surpassing} proved that models trained on synthetic data generalize well on real-world datasets. Following RandPerson, UnrealPerson \cite{zhang2021unrealperson} improved the accuracy by using real-world person images to create virtual characters, and rendering with the powerful Unreal Engine 4 (UE4) \cite{ue4engine} with four large and realistic scenes. Specifically, it cropped blocks from segmented clothing images to directly replace UV maps of existing 3D clothes models. However, as shown in Fig. \ref{fig:characters}, this way still results in unrealistic-looking characters due to scale alignment issue. Statistics of some synthetic datasets are shown in Table \ref{tab:dataset_compare}.

On the other hand, one may consider using virtual try-on methods to generate synthesized persons. These methods aim to transfer a target clothing onto a reference person. 
However, existing virtual try-on methods \cite{han2018viton, yang2020towards} are mostly in 2D, which cannot generate 3D clothed human models, and thus cannot import them into virtual environments for comprehensive rendering. 
On the other hand, some existing methods, e.g. PIFu \cite{saito2019pifu}, targets at high-fidelity reconstruction of 3D persons from 2D images. However, such methods require ground-truth of 3D shapes and textures for training, which is quite expensive and limited in scale.
Recently, some methods, e.g. HPBTT~\cite{zhao2020human} tried training 3D reconstruction models from only 2D images. However, they are based on generative models, which usually result in blurred textures and artifacts.
Besides, Pix2Surf \cite{mir2020learning} proposed to transfer texture from clothing images to 3D humans by neural networks. It achieved a good quality by training a specific model for every category of clothes. However, extending to other categories is costly. Furthermore, since HPBTT and Pix2Surf are both based on SMPL \cite{loper2015smpl}, they are not able to handle long skirts, as shown in Fig. \ref{fig:compare}.

Therefore, to further reduce the gap between virtual characters and realistic persons, we follow the way of RandPerson in repainting UV maps of existing 3D clothing models. However, different from RandPerson and UnrealPerson which directly replace existing UV maps by other images, we design two cloning methods for structure-aware, fine-grained repainting of UV maps.

\section{3D Virtual Character Generation}
\label{sec:pipeline}

\subsection{Pipeline Overview}

\cref{fig:pipeline_img} shows the pipeline of the proposed ClonedPerson approach, which includes the following steps. Firstly, we apply pre-processing steps, including pedestrian detection, pose detection, clothes detection, and clothes keypoint detection to get qualified frontal-view person images and obtain the clothes positions, categories, and clothes keypoints. Next, two cloning methods, registered mapping and homogeneous expansion, are applied to clone clothes from person images to UV texture maps and generate 3D characters. Finally, following RandPerson \cite{wang2020surpassing}, these characters are imported into Unity3D to render synthesized person images.

Several pre-processing steps are implemented to fulfill our target. For example, to clone the full-body outfits from real-world person images to virtual 3D characters, we apply person detection to localize full-body person images, and remove standalone and occluded clothes. Besides, we design a number of rules based on pose detection to cherry-pick\footnote{No need to worry about dropping other images including some images in good conditions, since there are huge available sources.} non-occluded frontal-view person images, since they best show clothing patterns and collocations.
Due to space limits, pre-processing steps are introduced in Appendix B.

%-------------------------------------------------------------------------

\subsection{Registered Clothes Mapping}
\label{sec:perspective}

\begin{figure}
  \centering
  \begin{subfigure}{0.8\linewidth}
    \includegraphics[width=1\linewidth]{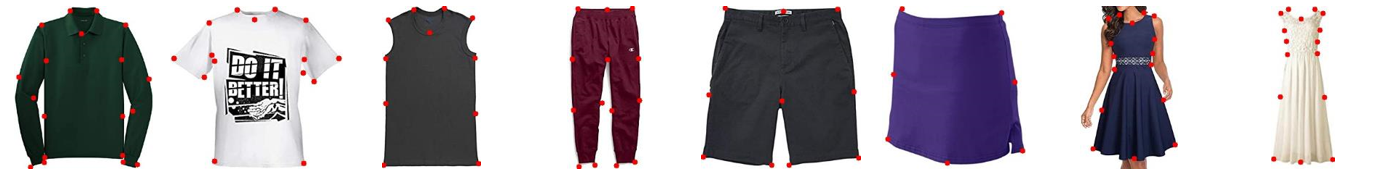}
    \caption{Eight clothes categories with labeled keypoints.}
    \label{fig:texture_a}
  \end{subfigure}
  \hfill
  \begin{subfigure}{0.8\linewidth}
    \includegraphics[width=1\linewidth]{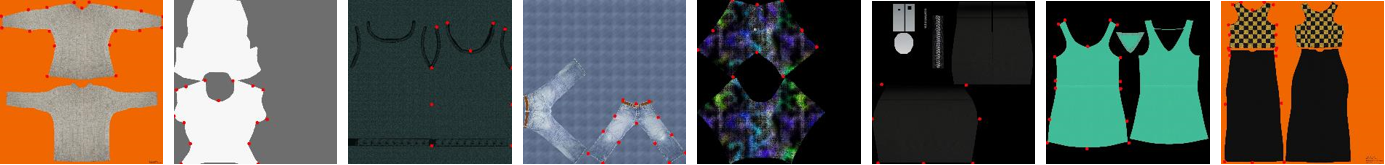}
    \caption{Regular UV maps where clothes appear in regular shapes and structures.}
    \label{fig:texture_b}
  \end{subfigure}
  \hfill
  \begin{subfigure}{0.8\linewidth}
    \includegraphics[width=1\linewidth]{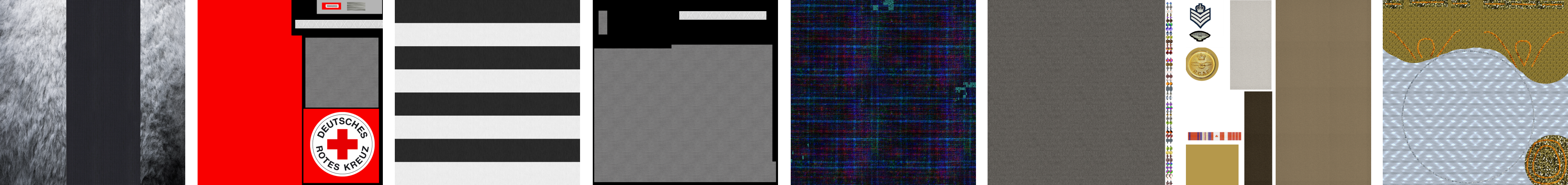}
    \caption{Irregular UV maps.}
    \label{fig:texture_c}
  \end{subfigure}
  \caption{Different categories of clothes and UV texture maps of the corresponding 3D clothes models. }
  \label{fig:textures}
\end{figure}

With 3D clothes models available in the MakeHuman community, we obtain some clothes models with regular UV maps, where clothes appear in regular shapes and structures, as \cref{fig:texture_b} shows. With these regular UV maps, we apply perspective homography \cite{szeliski2010computer} to map real-world clothes textures to UV maps of 3D characters, so that the original texture structures in the clothes can be well kept, and will appear to be clear and sharp.

\subsubsection{Perspective homography}

Perspective homography is also known as perspective transformation \cite{andrew2001multiple,szeliski2010computer}. Given a set of 2D points $\{\mathbf{p}_i\}$ and a corresponding set of points $\{\mathbf{p}'_i\}$, augmented with homogeneous coordinates (appending 1 as the $z$ coordinate), perspective homography maps each $\mathbf{p}_i$ to $\mathbf{p}'_i$ by a homography matrix $\mathbf{H}\in\boldsymbol{R}^{3\times3}$, that is, $\mathbf{p}' = \textbf{H}\mathbf{p}$.

Then, we can compute the homography matrix $\textbf{H}$ by solving the following optimization problem: 

\begin{align}
    \underset{\textbf{H}}{min}\sum_{i=1}^{n}\left \|  \mathbf{p}_i' - \textbf{H}\mathbf{p}_i \right \|_2^{2},
    \label{perspective_eq2}
\end{align} 
where $n$ is the number of the corresponding points. Eq. (\ref{perspective_eq2}) defines a least squares problem and thus can be easily solved. In addition, the computed homography matrix $\textbf{H}$ can be refined with the Levenberg-Marquardt method \cite{10.2307/43633451} to further reduce the re-projection error. 

\subsubsection{Perspective warping}

\begin{figure*}
  \centering
  \includegraphics[width=0.8\linewidth]{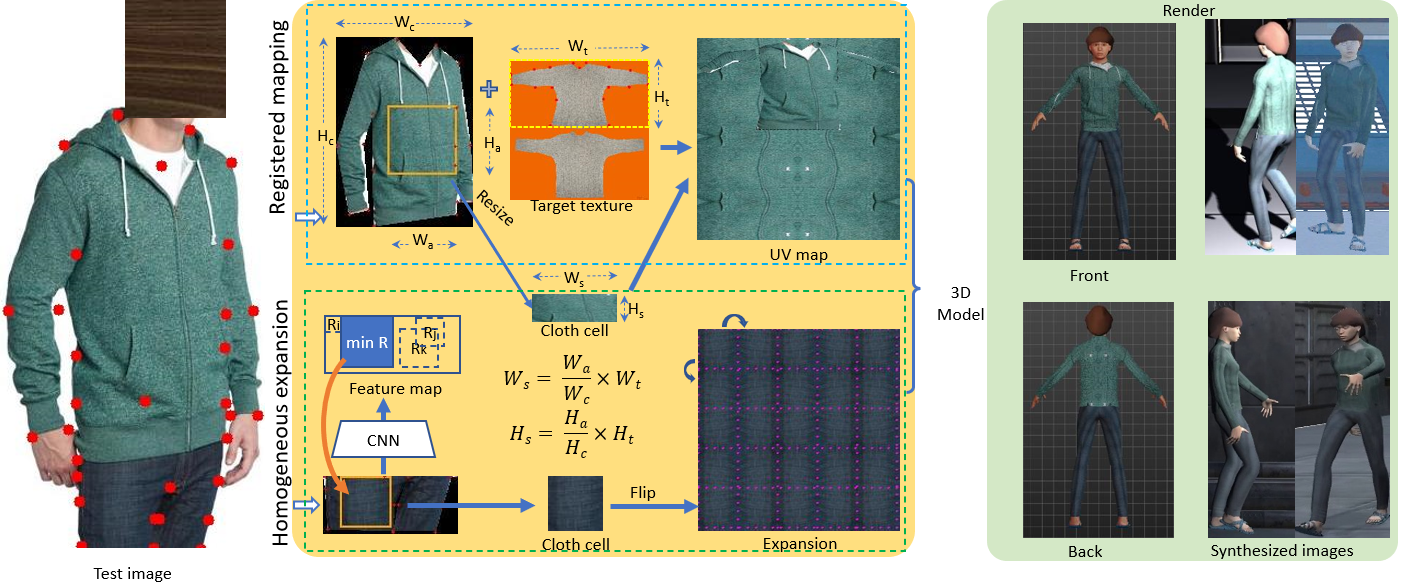}
  \caption{Process pipeline of registered clothes mapping (top) and homogeneous cloth expansion (bottom).}
  \label{fig:create_texture}
\end{figure*}

In our task, given labeled clothes keypoints $\mathbf{p}_i$ on regular UV texture maps (e.g. \cref{fig:texture_b}), and the corresponding detected clothes keypoints $\mathbf{p'}_i$ on real-world person images (e.g. \cref{fig:texture_a}), we can solve Eq. (\ref{perspective_eq2}) and get the homography matrix $\textbf{H}$. Then, each pixel location $\mathbf{p}$ on the UV map will have a corresponding pixel location on the input image, by $\left [ x, y, z \right ]^{T} = \textbf{H}\mathbf{p}$. Besides, we need to set the $z$ coordinate of all the resulting points to 1 before the warping process, as the transformation operates on homograph coordinates. That is,
$\mathbf{p'} =  \left (\frac{x}{z}, \frac{y}{z}  \right )$, where $\mathbf{p'}$ represents the corresponding point on the clothes image. Finally, the perspective warping can be done by bilinear interpolation on the clothes image and use the resulting pixel values to fill the UV map \cite{szeliski2010computer}. Specifically, by traversing $\mathbf{p}$ on UV map in turn, a corresponding point $\mathbf{p'}$=$\textbf{H}\mathbf{p}$ with float numbers of coordinates on the clothing image will be determined. Then, four pixels around $\mathbf{p'}$ will be bilinearly interpreted into $\mathbf{p}$. An example is shown in \cref{fig:create_texture}, where the red dots represent the corresponding keypoints. To reduce background influence, we set the outer part of the clothes in black.

In this paper, with most regular UV maps we directly calculate perspective homography through all the clothes keypoints. However, as shown in \cref{fig:texture_b}, the shapes of the long-sleeved and trousers on the UV map are quite different from those usually appear in person images. In these cases, we calculate the perspective homography on each part of them separately, then warp the clothes parts to the UV texture map and combine the results. 
For example, pants could be split into left and right sides.

\subsection{Homogeneous Cloth Expansion}
\label{subsec:textures}

Registered clothes mapping can handle the clothes texture of the frontal side very well. However, the back side is usually different from the frontal side but invisible. Therefore, we further design the homogeneous cloth expansion method to find a homogeneous area on clothes as a realistic cloth cell, and expand the cell to fill the UV map. Besides, as \cref{fig:texture_c} shows, the UV texture maps of some 3D clothes models are irregular, with unclear clothes structures. This also prevents the application of the registered clothes mapping. Therefore, we use the homogeneous cloth expansion to handle irregular UV maps, and enable more clothes models and styles. In our experiments, we have 161 3D clothes models with regular UV maps, and 17 models with irregular UV maps, as illustrated in \cref{fig:textures}.

\subsubsection{Cloth segmentation}

The homogeneous cloth expansion includes two steps, cloth segmentation and cloth expansion. For cloth segmentation, an optimization algorithm is proposed to find a large homogeneous area on clothes, and use it as a realistic cloth cell. As shown in \cref{fig:create_texture}, we first crop the clothes area, and use a model trained on MSMT17 \cite{wei2018person} by QAConv 2.0  \cite{Liao-2021-QAConv-GS} to extract the layer2 feature map ($48\times16$) of this clothes image. The purpose of the QAConv model is to extract discriminant feature maps to find homogeneous cells, so as to reduce the influence of image noises. Then, square blocks of various scales are defined on the feature map. Within each block, the average and standard deviation of the feature values are computed, as follows:

\begin{equation}
  \boldsymbol{\mu}^k = \frac{1}{n_k} \sum_{i=1}^{n_k}{\mathbf{x}_{i}^k},
  \hspace{3mm}
  {\sigma}_j^k = \sqrt{\frac{1}{n_k-1} \sum_{i=1}^{n_k}({x_{ij}^k - \mu_j^k)^2}},
\end{equation}
where $k$ denotes the $k$th block, $n_k$ is the number of elements in that block, $\mathbf{x}_{i}^k\in \boldsymbol{R}^d$ is the feature vector of the $i$th element in block $k$, with $d=512$ dimensions, and $j$ denotes the $j$th dimension. Note that the standard deviation is computed per feature channel. This value estimates the variations within each block, and thus reflects how homogeneous the cloth is within that block. Furthermore, we would also like the selected block to be as large as possible. Therefore, we further compute the area $A_k$ of each block $k$, and define a ratio $R$ as our objective function for the optimization problem, as follows:
\begin{equation}
  \min_{k=1}^K R_k = \frac{\frac{1}{d} \sum_{j=1}^{d} \sigma_j^k}{A_{k}},
  \label{eq:score}
\end{equation}
where $K$ denotes the number of blocks. By optimizing the above objective, we obtain a cloth area, with textures within it as homogeneous as possible, and with the area as large as possible. Then, we locate this block on the input clothes image and crop it, resulting in a patch which we call cloth cell. Appendix Fig. I shows some cloth cells thus obtained.
% Some cells thus obtained are shown in Appendix D.

\subsubsection{Cloth Expansion}

As described above, the homogeneous cloth expansion is applied for both regular UV maps and irregular UV maps. For regular UV maps, it is used to fill the back side of the clothes area, as well as the background. Since we already apply the registered clothes mapping for the frontal side of the clothes on regular UV maps, there exists a scale alignment problem for the cloth cell to be filled on the same UV map. Therefore, to maintain the consistency of the texture of the clothes, we need to scale the homogeneous cloth cell. As shown in \cref{fig:create_texture}, let $W_{c}$ and $H_{c}$ be the width and height, respectively, of the clothes image, $W_{a}$ and $H_{a}$ be the width and height, respectively, of the cropped cloth cell from the clothes image, $W_{t}$ and $H_{t}$ be the width and height, respectively, of the target area of the clothes after registered mapping, then,  $W_{s}$ and $H_{s}$, the width and height, respectively, of the cell to be scaled can be computed as follows:
\begin{equation}
  W_{s} = \frac{W_{a}}{W_{c}} \times W_{t},  \qquad
  H_{s} = \frac{H_{a}}{H_{c}} \times H_{t}.
  \label{eq:scale}
\end{equation}

Then, the scaled cloth cell is expanded on the whole UV map besides the registered clothes mapping area, by alternating flipping and tiling. As for irregular UV maps, since there is no reference of the scale, we simply use the original shape of the homogeneous cloth cell to flip and tile until fully fill the whole UV map, as shown in \cref{fig:create_texture}. 

Note that besides the homogeneous cloth expansion 
% method
, a simple way is to resize the cloth cell directly as a UV map, as done in RandPerson and UnrealPerson. However, simply resizing the cloth cells may result in blur textures and unrealistic patterns, as compared in Appendix D.

\section{Similarity-Diversity Expansion}\label{sec:similarity-diversity}

We use both DeepFashion (Apache License 2.0) \cite{liuLQWTcvpr16DeepFashion} and DeepFashion2 \cite{DeepFashion2} images for our virtual data creation. Through the pre-processing steps, there are still tens of thousands of 
% person and clothes 
images that are qualified and can be cloned to virtual characters. However, because of the enormous volume of images and repeating images of the same person, using these images directly to create characters is not efficient. To address this, two principles are considered. First, the more diverse samples, the better generalization performance should be. Second, similar person images are able to make the model training pay more attention to subtle differences. According to \cite{zhang2021unrealperson}, similar characters as \textit{hard samples} take a positive effect for person re-identification with large number of identities and cameras.

\begin{figure}
  \centering
  \includegraphics[width=0.8\linewidth]{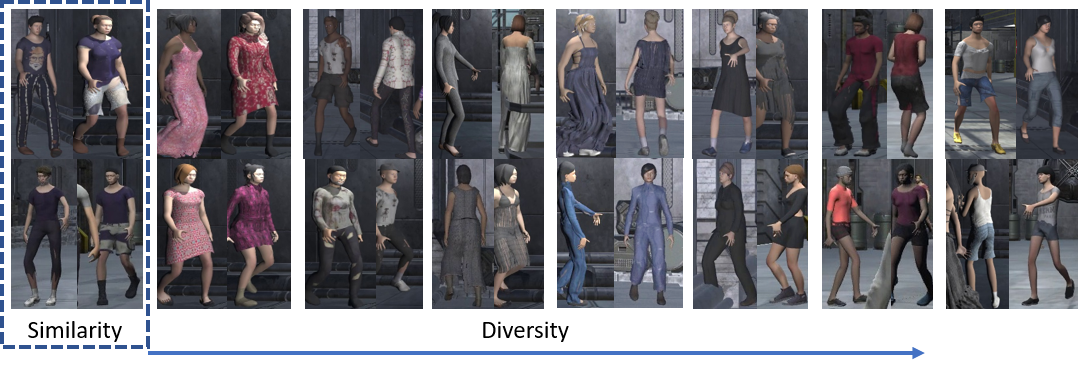}
  \caption{Illustration of similarity-diversity expansion.}
  \label{fig:similarity_diversity}
\end{figure}

\begin{figure}
  \centering
  \begin{subfigure}{0.25\linewidth}
    \includegraphics[width=1\linewidth]{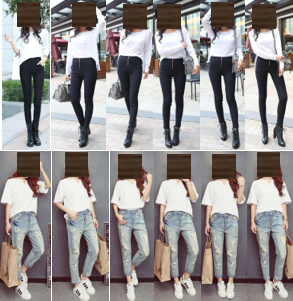}
    \caption{$\epsilon$ = 0.4}
    \label{fig:cluster_0.4}
  \end{subfigure}
  \hfill
  \begin{subfigure}{0.6\linewidth}
    \includegraphics[width=1\linewidth]{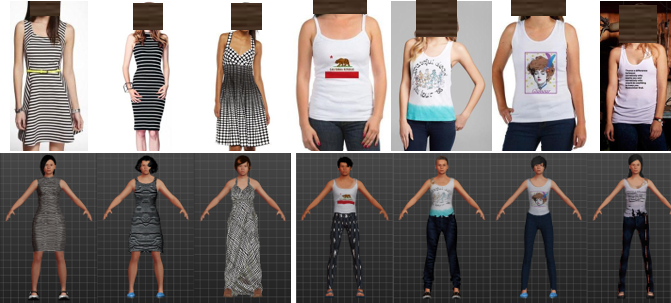}
    \caption{$\epsilon$ = 0.5}
    \label{fig:cluster_0.5}
  \end{subfigure}
  \caption{Clustering of different $\epsilon$ values. (a) Examples of two clusters (up and down) with $\epsilon$=0.4. (b) Examples of two clusters (left and right) with $\epsilon$=0.5, with person images (first row) and the generated characters (second row).}
  \label{fig:cluster_result}
\end{figure}

Therefore, we propose a similarity-diversity expansion strategy to scale up virtual character creation while improving along both the similarity and diversity aspects, as illustrated in \cref{fig:similarity_diversity}. 
% The basic idea is illustrated in \cref{fig:similarity_diversity}. 
By clustering person images, we can create similar characters from the same cluster, while increase diversity by including more and more clusters. This way, the created characters can expand densely in similarity and diversely in population. Specifically, this strategy first applies DBSCAN \cite{ester1996density} to cluster person images, then it samples a certain number of images per cluster, and finally clones outfits from these images for 3D character generation. In this way, we can generate similar characters in the same cluster and diverse characters with different clusters \footnote{Note that we only create one 3D character for one person image, since generating similar characters is already considered in the same cluster.}.

For the clustering, we use the same model trained on MSMT17 \cite{wei2018person} by QAConv 2.0 \cite{Liao-2021-QAConv-GS} to extract feature maps and compute similarity scores between person images. Then, DBSCAN is applied, with different $\epsilon$ parameters to control the degree of similarity. Specifically, to remove repeating persons, we set $\epsilon$=0.4 to cluster the same person with the same outfits. \cref{fig:cluster_0.4} shows two examples where images from the same cluster are with the same person. Next, we select one image per cluster (closest to the cluster center) and remove other redundant images. Together with other images failed to be clustered (with label -1), the second round of clustering is performed, with $\epsilon$=0.5. As shown in \cref{fig:cluster_0.5}, this time images from the same cluster are visually similar but generally from different persons.

Finally, we select seven  images (five for training and two for testing) per cluster to generate characters. Following RandPerson \cite{wang2020surpassing}, these characters are imported into Unity3D 
% virtual environments 
to render synthesized person images. We implement some adjustments to improve the rendering, as detailed in Appendix E. Accordingly, we create 5,621 characters with 887,766 images, as the ClonedPerson dataset, with 763,953 images from 4,826 characters for training, and 123,813 images from 795 characters for testing. The  Statistics of the dataset are shown in Table \ref{tab:dataset_compare}. We summarize the details and statistics of each step in our pipeline in Appendix F. \cref{fig:similarity_diversity} and \cref{fig:clonedperson} (more in the Appendix) show some characters created in ClonedPerson.

\section{EXPERIMENTS}

\subsection{Datasets}

Three real-world person re-identification datasets, CUHK03 \cite{li2014deepreid}, Market-1501 \cite{zheng2015scalable}, and MSMT17 \cite{wei2018person}, are used for generalization evaluation. The CUHK03 dataset includes 14,097 images of 1,467 individuals. There are 7,365 images of 767 identities in the training set, and 6,732 images of 700 identities in the test set, according to the CUHK03-NP protocol \cite{zhong2017re}. The detected bounding boxes are used. The Market-1501 dataset includes 32,668 images of 1,501 identities captured from six cameras. 12,936 images of 751 identities are included in the training set, and the remaining 19,732 images of 750 identities are used for the test set. The MSMT17 dataset includes 126,441 images of 4,101 identities and divided into 32,621 images of 1,041 identities for training, and the remaining 93,820 images of 3,060 identities for testing. 

We use two other synthetic datasets, RandPerson \cite{wang2020surpassing} and UnrealPerson \cite{zhang2021unrealperson}, for comparison, since they are shown to be superior to other synthetic datasets for generalizable person re-identification. RandPerson contains 8,000 identities and 1,801,816 images with 19 cameras. Both the full set and the suggested subset (132,145 images of the 8,000 identities) are used for our experiments. Besides, since some rendering setups are modified in this work, we further render RandPerson characters in our conditions for a fair comparison. This is denoted by RandPerson$^*$ (RP$^*$). UnrealPerson releases 6,799 characters with 1,256,381 images. Both the full set and the suggested subset (120,000 images of 3,000 identities) are used for our experiments.

\subsection{Methods}

The validation of the proposed ClonedPerson 
% dataset 
is through person re-identification experiments. We mainly consider two tasks, generalizable person re-identification \cite{hu2014cross,yi2014deep,liao2020interpretable}, and unsupervised domain adaptation (UDA). We apply QAConv 2.0 \cite{Liao-2021-QAConv-GS} and TransMatcher \cite{liao2021transmatcher} for the former, and SpCL \cite{ge2020selfpaced} for the latter. All of them are under the MIT License. We keep the same settings for each method.

All evaluations follow the single-query evaluation protocol \cite{farenzena2010person}. We use the Cumulative Matching Characteristic (CMC) \cite{Phillips-HFR2}, especially the Rank-1 accuracy, and the mean Average Precision (mAP) \cite{sobh2010innovations} as the performance metrics. 

\subsection{Generalizable Person Re-Identification}

The mAP results of direct cross-dataset evaluation are shown in Table \ref{tab:dataset_result} comparing real-world datasets with QAConv 2.0, and Table \ref{tab:difmethod_result} comparing synthetic datasets with QAConv 2.0 and TransMatcher. Rank-1 results are reported in Appendix Table C.
In overall, ClonedPerson achieves the best performance, surpassing existing datasets of both synthetic and real-world. The better performance over existing real-world datasets further confirms the findings in \cite{wang2020surpassing} and \cite{zhang2021unrealperson}. Besides, 
% it can be seen that, 
ClonedPerson is much better than UnrealPerson on CUHK03 and Market-1501, while they are comparable on MSMT17. Note that scenes used by UnrealPerson are more large and realistic than ours, due to the powerful UE4 engine. Besides, UnrealPerson has ten more cameras than ClonedPerson. 
Note also that, by comparing to both full set and subset results of RandPerson and UnrealPerson, it is clear that ClonedPerson's better performance is not because it is larger, but because of its capability of cloning the whole outfits from person images.

Moreover, by comparing RandPerson$^*$ to RandPerson, the new rendering settings are more effective. Besides, compared to RandPerson$^*$ with the same rendering setting, ClonedPerson has gained an averaged improvement of about 2\% in mAP. This is 
% quite
encouraging since ClonedPerson has only 4,826 identities, compared to 8,000 in RandPerson.

\begin{table}

 \begin{center}
  \begin{tabular}{@{}c@{}|@{}c@{}|@{}c@{}|@{}c@{}|@{}c@{}|@{}c@{}|@{}c@{}}
   \hline
     \multirow{2}{*}{Dataset}& \multicolumn{2}{c|}{CUHK03-NP} & \multicolumn{2}{c|}{Market-1501} &  \multicolumn{2}{c}{MSMT17}  \\
    \cline{2-7}
        & Rank-1 & mAP & Rank-1 & mAP & Rank-1 & mAP \\
   \hline\hline
   CUHK03    & -     & -     & 65.5  & 34.5  & 42.3  & 13.4 \\
   Market-1501  & 15.1  &15.1   & -     & -     & 41.0  & 14.6    \\
   MSMT17       & 18.5  &19.2   &76.3   & 47.9  & -     & -     \\
   \hline\hline
%   \hline\hline
    ClonedPerson &\textbf{22.6}&\textbf{21.8}&\textbf{84.5}&\textbf{59.9}& \textbf{49.1} &  \textbf{18.5} \\
   \hline
  \end{tabular}
   \caption{Direct cross-dataset evaluation results of models trained on different datasets with QAConv 2.0. 
   }
   \label{tab:dataset_result}
 \end{center}
\end{table}

\begin{table}

 \begin{center}
  \begin{tabular}{c|c|c|c|c}
   \hline
     \multirow{2}{*}{Training Data}& \multicolumn{2}{c|}{QAConv} & \multicolumn{2}{c}{SpCL}  \\
    \cline{2-5}
        & Rank-1 & mAP & Rank-1 & mAP \\
   \hline\hline
   CUHK03 &\cellcolor{lime}29.1 &\cellcolor{lime}2.8&\cellcolor{cyan} 11.5  & \cellcolor{cyan}1.0  \\
   Market-1501  & \cellcolor{lime}40.3  &\cellcolor{lime}5.9   &\cellcolor{cyan} 12.0     &\cellcolor{cyan} 1.1     \\
   MSMT17       & \cellcolor{lime}39.8  &\cellcolor{lime}6.3   &\cellcolor{cyan}10.2   &\cellcolor{cyan} 0.9  \\
   \hline
    ClonedPerson &\cellcolor{lightgray}91.1&\cellcolor{lightgray}68.9&\cellcolor{pink}10.6&\cellcolor{pink}0.9\\
   \hline
  \end{tabular}
   \caption{Evaluation results on the ClonedPerson testing set with different tasks. Green: Cross-dataset evaluation. Gray: Within-dataset evaluation. Blue: UDA. Pink: Unsupervised Learning.
   }
   \label{tab:clonedperson_result}
 \end{center}
\end{table}

Furthermore, with the same learned QAConv models in Tables \ref{tab:dataset_result} and \ref{tab:difmethod_result}, we also evaluate their performances on the ClonedPerson testing set, with results shown in Table \ref{tab:clonedperson_result}. First, with within-dataset evaluation, QAConv achieves 91.1\% in Rank-1 and 68.9\% in mAP, indicating that this synthetic domain can be reasonably fitted by a representative 
method. Second, with cross-dataset evaluation, it can be seen that all models trained on real-world datasets perform not satisfactory on ClonedPerson, indicating that ClonedPerson is quite different and challenging. Nevertheless, it can still be observed that MSMT17 and Market-1501 are more diverse for generalization than CUHK03.

\begin{table}
 \begin{center}
  \begin{tabular}{@{}c@{}|@{}c@{}|c|c|@{}c@{}|@{}c@{}|@{}c@{}|@{}c@{}}
   \hline
     Method&Dataset& \#ID & \#Imgs& CUHK & Market & MSMT& Avg  \\

  \hline\hline
  \multirow{6}*{QAConv}&$RP$& 8,000 & 1,801k& 16.0 & 46.9 & 14.0&25.6\\
     ~& $RP$ &8,000& 132k & 15.1   &45.9     &13.8  &  24.9   \\
        ~&$RP^*$ & 8,000& 1,239k & 20.1  &56.4   &17.6  & 31.4     \\
   ~ &$UP$ & 6,799&1,256k  & 17.2 & 56.1 & 17.5&30.3 \\
  ~ &$UP$ &3,000&120k&17.8  &55.9  &\textbf{19.3} &31.0\\
  ~ &$CP$ & 4,826 & 763k &\textbf{21.8} & \textbf{59.9}&18.5&\textbf{33.4}\\
  \hline\hline
  
  \multirow{6}*{TransM} & $RP$&8,000 & 1,801k& 18.7 & 49.6&16.4&28.2\\
  ~&$RP$&8,000 & 132k& 16.9& 49.0&15.8&27.2\\
  ~&$RP^*$&8,000 & 1,239k& 22.9& 58.0&20.9&33.9\\
  ~ & $UP$ & 6,799&1,256k  & 19.7 & 60.2&18.4&32.8 \\
  ~ & $UP$ & 3,000&120k  & 19.6 & 59.4&\textbf{21.6}&33.5 \\
  ~ & $CP$& 4,826 & 763k & \textbf{24.4} & \textbf{62.3}&20.8&\textbf{35.8}\\
   \hline\hline
%   Msmt(SpCL) & 1,041 &30,248&  && \\
  \multirow{3}*{SpCL} &$RP$& 8,000 & 132k&4.7  &67.2&27.2 &33.0\\
    ~ &$UP$ & 3,000&120k  &5.3 &71.7 &\textbf{28.4} &35.1\\
  ~ &$CP$ & 4,826 & 75k &\textbf{ 12.0 }& \textbf{72.7}&24.2&\textbf{36.3}\\
   \hline
  \end{tabular}
   \caption{mAP results with different datasets for different tasks. TransM: TransMatcher. $RP$: RandPerson. $RP^*$: RandPerson$^*$ in new rendering settings. $UP$: UnrealPerson. $CP$: ClonedPerson. }% means an adapted RandPerson dataset rendered with the same settings of the ClonedPerson.}}
   \label{tab:difmethod_result}
 \end{center}
\end{table}

\subsection{Unsupervised Domain Adaptation}

As for UDA, we conducted experiments with SpCL, using ClonedPerson as source training data or target testing data. Since the whole training set of ClonedPerson is too large for SpCL to handle, e.g. in its clustering stage, we also selected a training subset of ClonedPerson for SpCL, with one image per camera, and 75,830 images in total from the 4,826 subjects. 
With ClonedPerson as source training data, the results are shown in Table \ref{tab:difmethod_result}. It 
% can be observed 
shows that on average ClonedPerson outperforms both RandPerson and UnrealPerson, especially, with a large margin on CUHK03. 

With ClonedPerson as target dataset, the results are shown in Table \ref{tab:clonedperson_result}. Similar as cross-dataset evaulation, the UDA results on ClonedPerson are also poor. Furthermore, we also conduct an unsupervised learning task on ClonedPerson by SpCL, as shown in Table \ref{tab:clonedperson_result}. Again, the results are poor. Therefore, it appears that, for SpCL, real-world source training data does not help 
% too 
much in domain adaptation to ClonedPerson, and thus the poor performance is mainly due to the unique challenges in ClonedPerson for clustering-based identity label reasoning. For example, there are a large number of diverse cameras and a lot of similar persons created by the proposed similarity-diversity expansion strategy. Consequently, though ClonedPerson is a synthesized dataset, it may provide a good test bed for both domain generalization and domain adaptation, and challenge researchers in developing more effective algorithms.

\subsection{Comparison of Different Generation Settings}

\begin{figure}
  \centering
  \begin{subfigure}{0.49\linewidth}
    \includegraphics[width=1\linewidth]{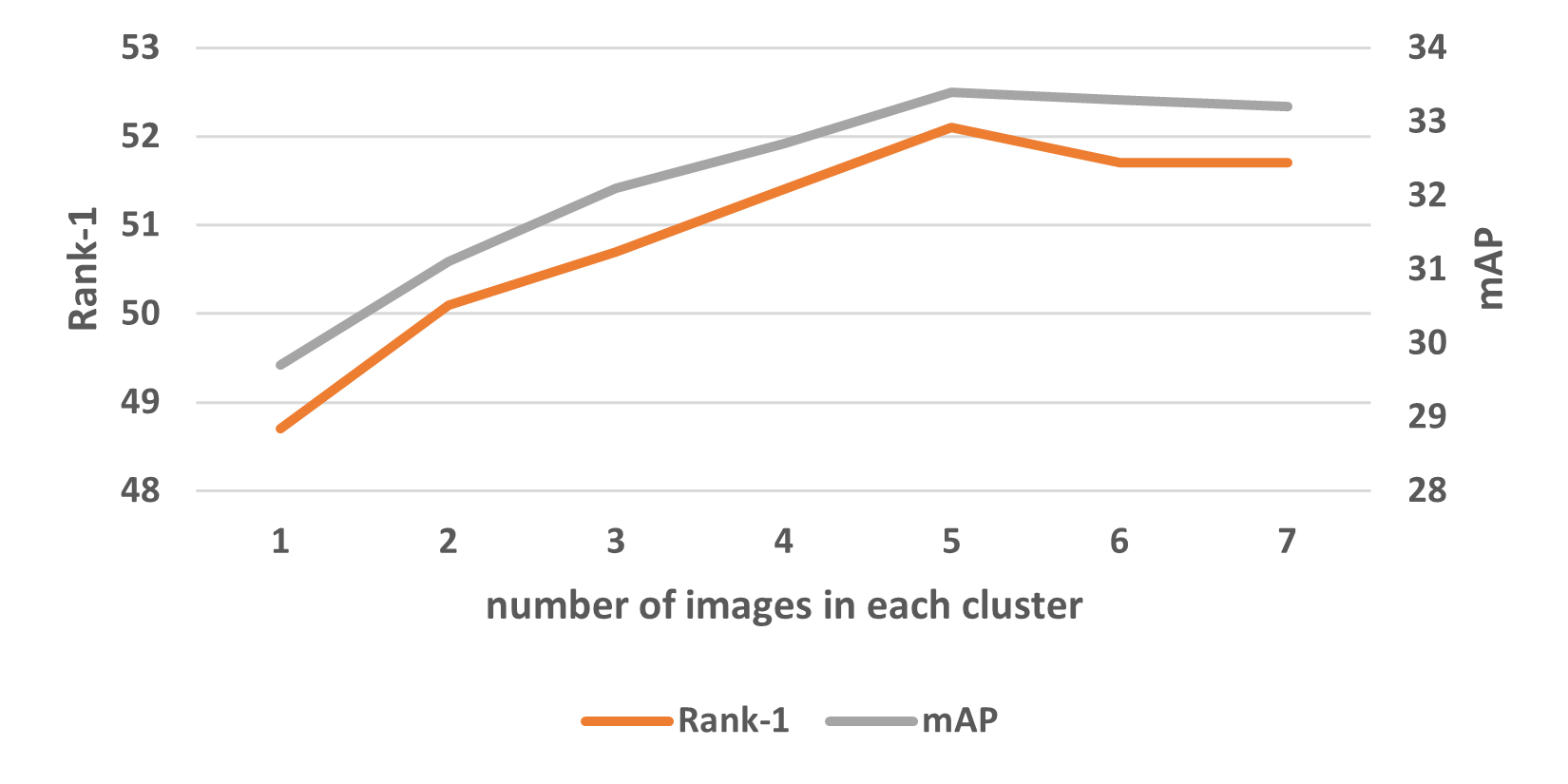}
    \captionsetup{font={scriptsize}}
    \caption{Different number of images per cluster}
    \label{fig:subset}
  \end{subfigure}
  \hfill
  \begin{subfigure}{0.49\linewidth}
    \includegraphics[width=1\linewidth]{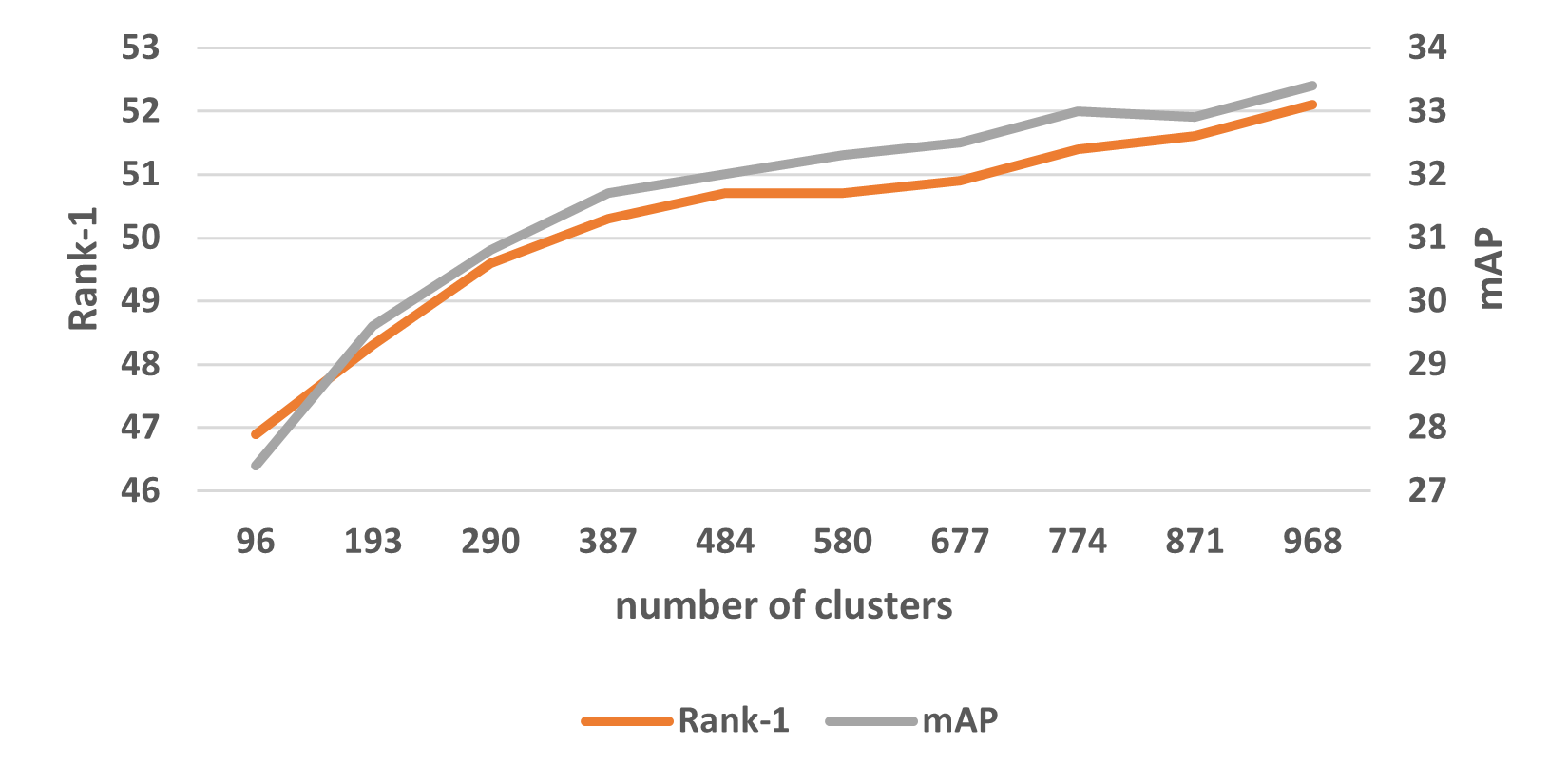}
    \captionsetup{font={scriptsize}}
    \caption{Different number of clusters}
    \label{fig:subclass}
  \end{subfigure}
  \hfill
  \begin{subfigure}{0.49\linewidth}
    \includegraphics[width=1\linewidth]{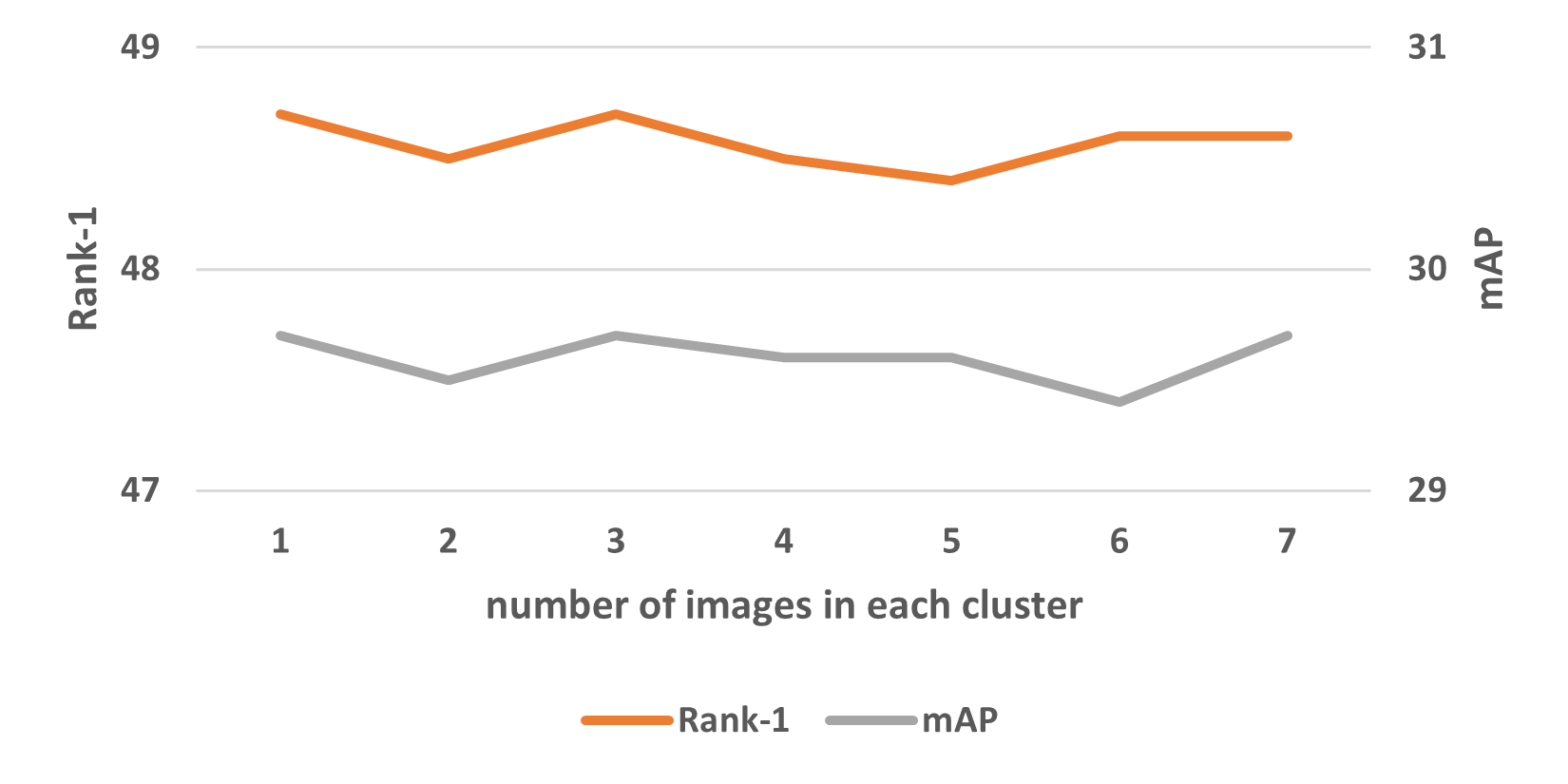}
    \captionsetup{font={scriptsize}}
    \caption{Same number of identities}
    \label{fig:sameid}
  \end{subfigure}
  \hfill
  \begin{subfigure}{0.49\linewidth}
    \includegraphics[width=1\linewidth]{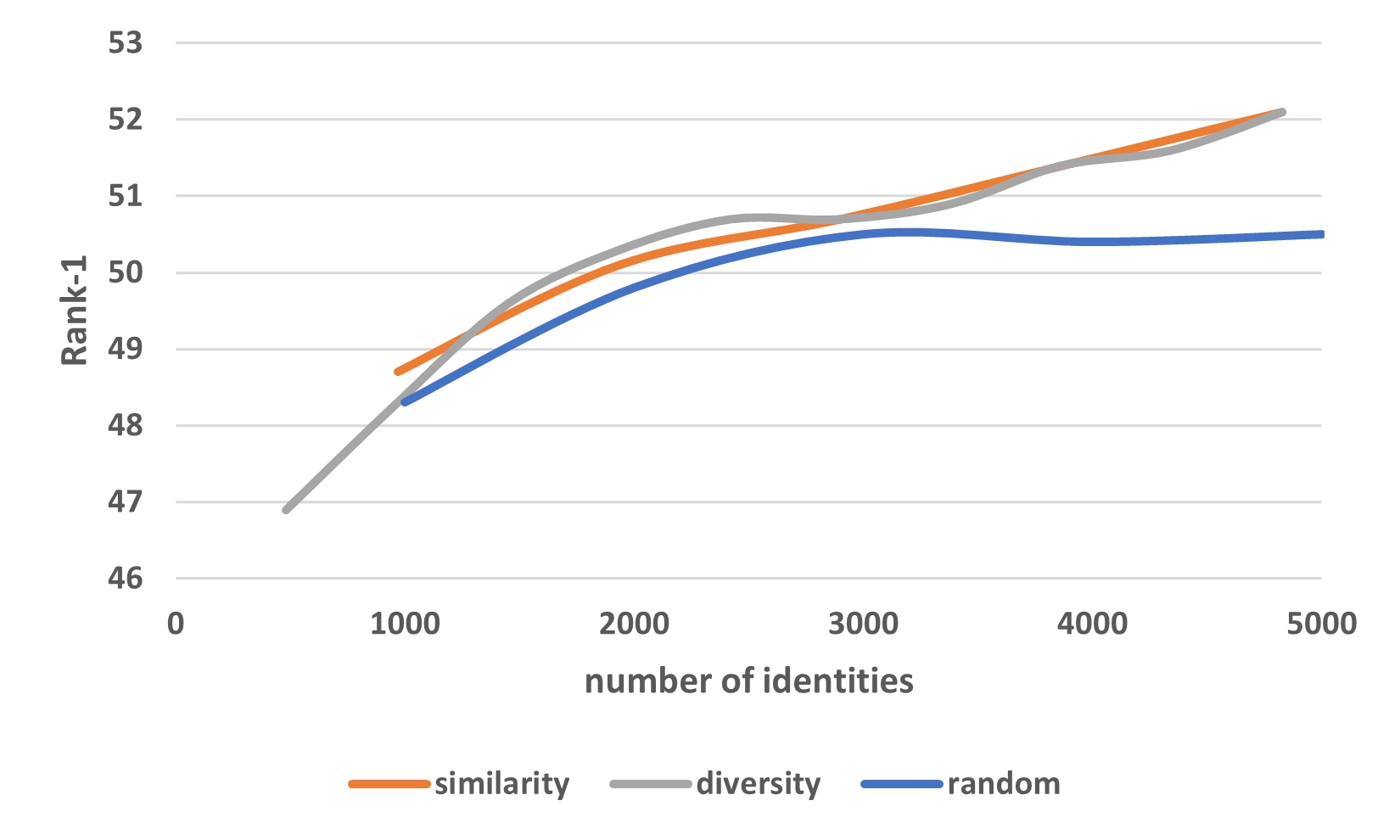}
    \captionsetup{font={scriptsize}}
    \caption{Different number of identities}
    \label{fig:random}
  \end{subfigure}
  \caption{Performance of different character scaling up methods. }
  \label{fig:performance_result}
\end{figure}

\cref{fig:performance_result} shows performance (averaged Rank-1 and mAP of the three real-world testing datasets) with different character scaling up methods, including different settings of the proposed similarity-diversity expansion strategy, and a straightforward random scaling up strategy. 

After the clustering procedure described in \cref{sec:similarity-diversity}, we obtain 968 clusters. Then, first, we use all the clusters for maximum diversity, and select different numbers of images per cluster for experiments, indicating increasing similarity. \cref{fig:subset} shows the performance. As the number of selected images increases, the performance clearly increases. Therefore, it proves that creating similar persons is indeed important for discriminant model learning, since it has to pay more attention to fine details of characters. However, it is saturating when the number of images per cluster reaches five. Therefore, to avoid data redundancy and improve efficiency, we select five images per cluster for the training set, and treat the remaining as a separate testing set.

Next, we keep the similarity level consistent, with five images per cluster, and select different numbers of clusters for experiments, indicating increasing diversity. As \cref{fig:subclass} shows, the performance increases as the number of clusters increases, which aligns with our expectations that performance raises as the diversity increases. 

However, the adjustment of diversity and similarity will inevitably cause changes in identities, which might influence performance. Therefore,  
% in addition, 
we keep the number of identities consistent by balancing the variation of similarity and diversity. That is, the selected clusters are gradually reduced when the number of images per cluster increases. The results are shown in \cref{fig:sameid}. 
% As can be seen, 
The performance fluctuates in a small range within 0.3\%, indicating that both similarity and diversity are important in our virtual data creation.

Finally, we also compare the random creation method that randomly selects person images for texture cloning and character creation before our clustering step. \cref{fig:random} shows the comparison of this random creation method to our strategies in \cref{fig:subset} and \cref{fig:subclass}. From the results it is clear that with random creation after 3,000 characters the performance is saturating. In contrast, the proposed similarity-diversity expansion strategy is much more efficient in scaling up the virtual character creation, especially with larger number of identities.

Therefore, the above analysis proved that the 
similarity-diversity expansion is effective and efficient in scaling up the virtual character creation, and is potentially useful in creating an even larger and effective dataset when more person image sources are considered, considering the trend in \cref{fig:subclass}. In contrast, in UnrealPerson \cite{zhang2021unrealperson} the conclusion is that it can only achieve the best performance with 3,000 characters, but more characters do not help. This is also verified in our experiments with UnrealPerson in \cref{tab:difmethod_result}.

\subsection{Qualitative Comparisons}

\begin{figure}
  \centering
  \begin{subfigure}{0.3\linewidth}
    \includegraphics[width=0.95\linewidth]{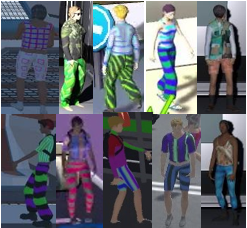}
    \caption{RandPerson}
    \label{fig:randperson}
  \end{subfigure}
  \hfill
  \begin{subfigure}{0.3\linewidth}
    \includegraphics[width=0.95\linewidth]{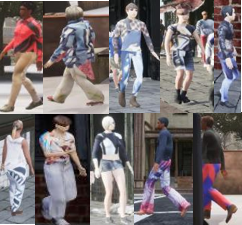}
    \caption{UnrealPerson}
    \label{fig:unrealperson}
  \end{subfigure}
  \hfill
  \begin{subfigure}{0.3\linewidth}
    \includegraphics[width=0.95\linewidth]{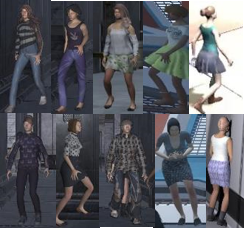}
    \caption{ClonedPerson}
    \label{fig:clonedperson}
  \end{subfigure}
  \caption{Examples from different synthetic datasets.}
  \label{fig:characters}
\end{figure}

\cref{fig:characters} shows some images of characters created by three different methods, RandPerson, UnrealPerson, and the proposed ClonedPerson. As can be seen, RandPerson is the most cartoon-like. As for UnrealPerson, though it also uses real clothes textures, most of its created characters do not match real-life clothes due to the scale alignment issue of cloth patterns. In contrast, thanks to the designed cloning pipeline, the ClonedPerson characters are more realistic and dressed more like real-life persons.

\begin{figure}
  \centering
  \begin{subfigure}{0.165\linewidth}
    \includegraphics[width=0.95\linewidth]{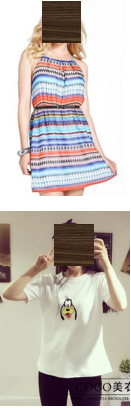}
    \captionsetup{font={scriptsize}}
    \caption{Test image}
    \label{fig:com_test}
  \end{subfigure}
  \hfill
  \begin{subfigure}{0.32\linewidth}
    \includegraphics[width=0.95\linewidth]{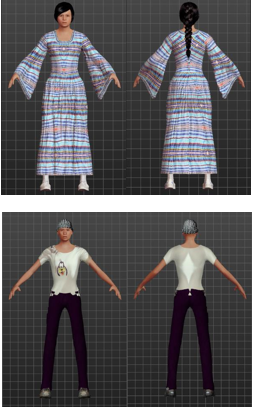}
    \captionsetup{font={scriptsize}}
    \caption{ClonedPerson}
    \label{fig:com_ours}
  \end{subfigure}
  \hfill
  \begin{subfigure}{0.2\linewidth}
    \includegraphics[width=0.95\linewidth]{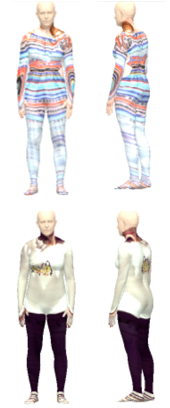}
    \captionsetup{font={scriptsize}}
    \caption{HPBTT}
    \label{fig:com_hpbtt}
  \end{subfigure}
  \caption{Qualitative comparison of different synthesis methods. }
  \label{fig:compare}
\end{figure}

Furthermore, \cref{fig:compare} shows a qualitative comparison between models created by our method and HPBTT \cite{zhao2020human}. It can be observed that characters created by the proposed method have clearer and sharper clothes textures, and better back-view looking of the clothes, than that generated by HPBTT. Besides, from the results shown in the first row, it can be seen that in ClonedPerson the clothes category is preserved, while HPBTT fails to deal with long skirts.

\section{CONCLUSION}

This paper contributes an automatic approach to clone the whole outfits from real-world person images to virtual 3D characters. Two critical cloning methods are proposed, registered clothes mapping and homogeneous cloth expansion. As a result, these characters bridge the gap between synthesized and realistic persons, and so models trained by our synthesized persons have better generalization ability for person re-identification. In addition, a similarity-diversity expansion strategy is proposed to scale up virtual characters. We show that similarity can help improve model's discrimination, while diversity can improve the generalization ability of the model. In the future, we could exploit more in developing different types of clothes models and exploit more data sources. Moreover, we show some limitations of this research in the Appendix.

%------------------------------------------------------------------------

%%%%%%%%% REFERENCES
{\small
\bibliographystyle{ieee_fullname}
\bibliography{egbib}
}
\clearpage
\include{Appendix}

\end{document}

%% file: Appendix.tex
% CVPR 2022 Paper Template
% based on the CVPR template provided by Ming-Ming Cheng (https://github.com/MCG-NKU/CVPR_Template)
% modified and extended by Stefan Roth (stefan.roth@NOSPAMtu-darmstadt.de)

% \documentclass[10pt,twocolumn,letterpaper]{article}

% % %%%%%%%%% PAPER TYPE  - PLEASE UPDATE FOR FINAL VERSION
% % % \usepackage[review]{cvpr}      % To produce the REVIEW version
% \usepackage{cvpr}              % To produce the CAMERA-READY version
% % %\usepackage[pagenumbers]{cvpr} % To force page numbers, e.g. for an arXiv version

% % % Include other packages here, before hyperref.

% \usepackage{graphicx}
% \usepackage{amsmath}
% \usepackage{amssymb}
% \usepackage{booktabs}
% \usepackage{multirow}
% \usepackage{ulem}
% \usepackage[labelformat=simple]{subcaption}
% \usepackage{tabularx}
% \usepackage{makecell}

% % It is strongly recommended to use hyperref, especially for the review version.
% % hyperref with option pagebackref eases the reviewers' job.
% % Please disable hyperref *only* if you encounter grave issues, e.g. with the
% % file validation for the camera-ready version.
% %
% % If you comment hyperref and then uncomment it, you should delete
% % ReviewTempalte.aux before re-running LaTeX.
% % (Or just hit 'q' on the first LaTeX run, let it finish, and you
% %  should be clear).
% \usepackage[pagebackref,breaklinks,colorlinks]{hyperref}

% Support for easy cross-referencing
%\usepackage[capitalize]{cleveref}
\crefname{section}{Sec.}{Secs.}
\Crefname{section}{Section}{Sections}
\Crefname{table}{Table}{Tables}
\crefname{table}{Tab.}{Tabs.}

% If you wish to avoid re-using figure, table, and equation numbers from
% the main paper, please uncomment the following and change the numbers
% appropriately.
%\setcounter{figure}{2}
%\setcounter{table}{1}
%\setcounter{equation}{2}

% If you wish to avoid re-using reference numbers from the main paper,
% please uncomment the following and change the counter for `enumiv' to
% the number of references you have in the main paper (here, 6).
% \let\oldthebibliography=\thebibliography
% \let\oldendthebibliography=\endthebibliography
% \renewenvironment{thebibliography}[1]{%
%     \oldthebibliography{#1}%
%     \setcounter{enumiv}{6}%
% }{\oldendthebibliography}

%%%%%%%%% PAPER ID  - PLEASE UPDATE
% \def\cvprPaperID{628} % *** Enter the CVPR Paper ID here
% \def\confName{CVPR}
% \def\confYear{2022}

% \begin{appendices}

% \maketitle
% %%%%%%%%% TITLE - PLEASE UPDATE
% \title{Cloning Outfits from Real-World Images to 3D Characters\\for Generalizable Person Re-Identification: Appendix}
\appendix
\twocolumn[
\begin{@twocolumnfalse}
    \section*{\centering{Cloning Outfits from Real-World Images to 3D Characters\\for Generalizable Person Re-Identification: Appendix\\[25pt]}}
\end{@twocolumnfalse}
]

% \author{Yanan Wang^{1}\\
% {\tt\small yanan.wang.cs@gmail.com}
% \and
% Xuezhi Liang^{2}\\
% {\tt\small xz.liang.cs@gmail.com}
% \and
% Shengcai Liao^{1} \thanks{Shengcai Liao is the Corresponding Author.}\\ 
% {\tt\small scliao@ieee.org}
% \and
% Inception Institute of Artificial Intelligence (IIAI)^{1}\\
% Mohamed bin Zayed University of Artificial Intelligence^{2}\\
% Masdar City, Abu Dhabi, UAE\\
% }

\setcounter{table}{0}
\setcounter{figure}{0}
\renewcommand{\thefigure}{\Alph{figure}}
\renewcommand{\thetable}{\Alph{table}}

\renewcommand\thesubfigure{(\arabic{subfigure})}

%%%%%%%%% BODY TEXT - ENTER YOUR RESPONSE BELOW
\section{Introduction}
Due to space limits, we are not able to explain everything in detail in the main paper. In this Appendix, we further present more details of our implementations, and demonstrate more illustrations to explain our design choices. Besides, we present more experimental results for further understanding. 
% We also make a demonstration video of the whole process and put it in the supplementary material.

All methods used and designed in the project are listed in \cref{tab:methods}, including existing methods, adapted methods, and the proposed methods. For example, we adapted some existing methods in the person image pre-processing stage to help cherry-pick best-viewing person images and determine clothes positions, categories and clothes keypoint locations. At the same time, we also propose some new methods, such as Registered Clothes Mapping, Homogeneous cloth Expansion and Similarity-Diversity Expansion, to achieve the goal of mapping real clothes to virtual people. Finally, we create the ClonedPerson dataset that can improve the generalization performance of person re-identification.

\begin{table*}[ht] 
 \begin{center}
  \renewcommand{\tabularxcolumn}[1]{m{#1}}
  \begin{tabularx}{\textwidth}{c|c|X}
   \hline
        \textbf{Method} & \textbf{Category} & \makecell{\textbf{Notes}} \\
   \hline
    Person Detection&	Existing	 &   Pedestron \cite{hasan2021generalizable}\\
   \hline
    Pose Detection&	Adapted	 &  We used the existing HRNet \cite{Sun_2019_CVPR_HRNET} model from MMDetection \cite{mmdetection}. Specific rules are designed based on the detected keypoints to cherry-pick best-viewing person images. \\
   \hline
    Clothes Detection and Classification & Adapted	 & We trained a model based on Faster-RCNN \cite{ren2015faster} with the annotated clothes bounding boxes and categories in DeepFashion2 \cite{DeepFashion2}. \\
   \hline
    Clothes Keypoint Detection & Adapted	 & We annotated clothes keypoints and trained a model based on PIPNet \cite{JLS21}. \\
   \hline
    Registered Clothes Mapping & Proposed & We annotated clothes keypoints on regular UV maps, detected clothes keypoints on person images, and applied the perspective homography method to warp real clothes texture to UV maps.\\
   \hline
    Homogeneous Cloth Expansion & Proposed  & A new method is proposed to find a homogeneous area as large as possible on clothes images. \\
   \hline
    Similarity-Diversity Expansion & Proposed & A new method is proposed to scale up virtual character creation.\\
   \hline
  \end{tabularx}
   \caption{Methods used and designed in the ClonedPerson pipeline.}
   \label{tab:methods}
 \end{center}
\end{table*}

\section{Person Image Pre-Processing}
\subsection{Pedestrian detection}
\label{subsec:pedestrian}

\begin{figure*}
  \centering
  \includegraphics[width=1\linewidth]{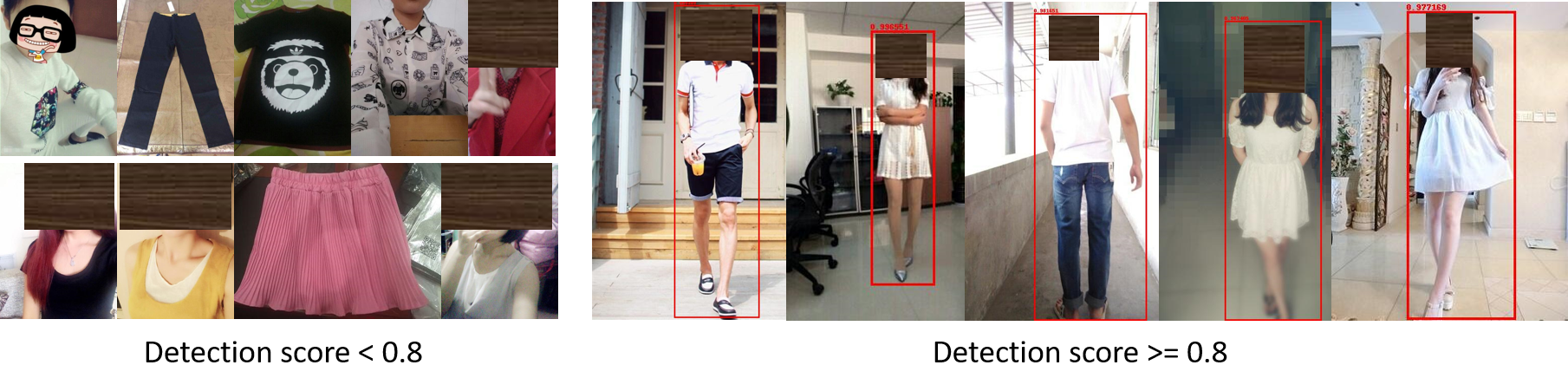}
  \caption{Examples of person detection results.}
  \label{fig:pede_result}
\end{figure*}

Our target is to clone the full-body outfits from real-world person images to virtual 3D characters. However, considering the variety of clothing images in real life, we need to avoid images of standalone clothes and incomplete person images. Therefore, we apply a person detection model, Pedestron \cite{hasan2021generalizable}, to detect qualified person images. We keep the original configuration of the Pedestron \cite{hasan2021generalizable} and set the detection threshold to be 0.8 to avoid images of standalone clothes and incomplete person images. Different situations of person images are shown in \cref{fig:pede_result}. Furthermore, we set the area of the detected bounding boxes to be at least 20\% of the input image to remove low-resolution persons and some false positives. The detected person images are cropped for the following pose detection procedure. 

%We use Pedestron \cite{hasan2021generalizable} to detect person images, and according to the detection score, by setting the detection threshold as 0.8, we could avoid images of standalone clothes and incomplete person images.

\begin{figure*}
  \centering
  \includegraphics[width=1\linewidth]{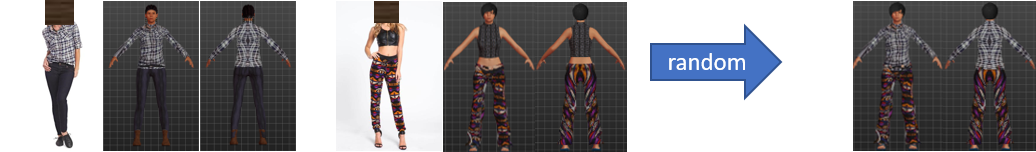}
  \caption{Examples of different combinations of upper-body and lower-body clothes. Left: the proposed cloning of the full-body outfits, in their original collocations. Right: random combination.}
  \label{fig:random_combine}
\end{figure*}

% In addition to avoiding standalone and occluded clothes, another reason for applying person detection is that our target is to clone the full-body outfits from real-world person images to virtual 3D characters. 
Characters in some existing synthetic datasets are dressed in random collocation, such as in RandPerson \cite{wang2020surpassing} and UnrealPerson \cite{zhang2021unrealperson}. However, random collocation sometimes creates incongruous characters, as shown in \cref{fig:random_combine}. The left side shows person images and characters created by the proposed cloning method. The right side shows characters created by randomly combining some upper-body and lower-body clothes. We can see that the collocation on the right is inconsistent. Therefore, the use of person detection in localizing full-body person images is also for the purpose of cloning the full-body outfits from real-world person images to virtual 3D characters. As a result, the proposed method follows the original collocations of real-life persons, and so the sample distributions of our data would be more consistent with real-life persons. 

\subsection{Person view qualification by pose detection}
\label{subsec:pose}

\begin{figure*}
  \centering
  \begin{subfigure}{0.2\linewidth}
    \includegraphics[width=1\linewidth]{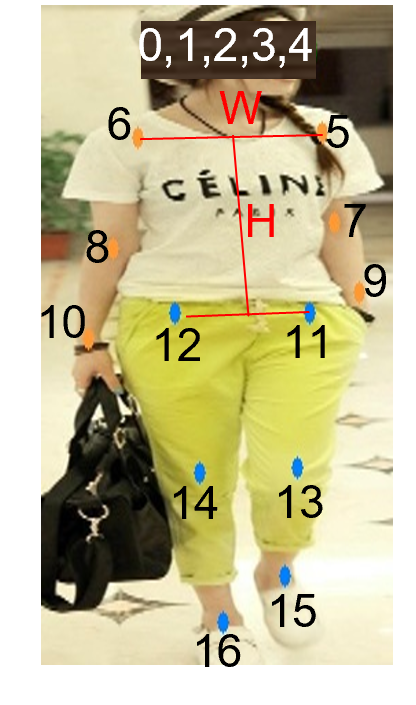}
    \caption{Qualified.}
    \label{fig:pose_a}
  \end{subfigure}
  \hspace{5mm}
  \begin{subfigure}{0.2\linewidth}
    \includegraphics[width=1\linewidth]{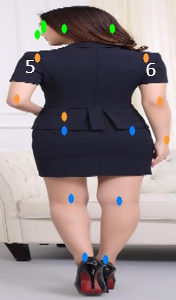}
    \caption{Back-view.}
    \label{fig:pose_b}
  \end{subfigure}
  \hspace{5mm}
  \begin{subfigure}{0.2\linewidth}
    \includegraphics[width=1\linewidth]{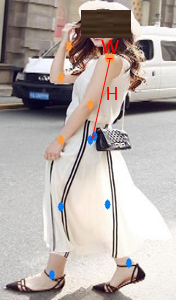}
    \caption{Side-view.}
    \label{fig:pose_c}
  \end{subfigure}
  \hspace{5mm}
  \begin{subfigure}{0.2\linewidth}
    \includegraphics[width=1\linewidth]{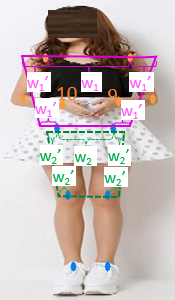}
    \caption{Occluded.}
    \label{fig:pose_d}
  \end{subfigure}
  \caption{Different viewpoints judged by pose detection. (1) A qualified image, where the left shoulder $P5$ is on the right side of the image, $W/H\ge0.3$, and hands are not in the body area. (2) A back-view image, where the left shoulder $P5$ is on the left side of the image. (3) A side-view image, where $W/H<0.3$. (4) An occluded image, with hands in the body area.}
  \label{fig:pose_result}
\end{figure*}

After person detection, another problem is that person images may have different viewpoints, such as frontal view, back view, and side view. Furthermore, the frontal view images are divided into two situations: occluded and non-occluded. For our purpose, back-view, side-view, and occluded front-view images are all incomplete displays of clothes, so they are regarded as noisy data. To reduce these noisy data, we use person pose estimation model for automatic judgment. Specifically, we apply the HRNet \cite{Sun_2019_CVPR_HRNET} model from MMDetection \cite{mmdetection} to do person pose estimation. It is trained on the COCO dataset \cite{lin2014microsoft}. HRNet predicts 17 body keypoints and their visibility probabilities, from which we use 12 keypoints on the body, including shoulders, elbows, hands, hips, knees, and feet. According to the positions of the shoulders, back-view images could be classified. With the width-to-height aspect ratio of the upper body, side-view images could be distinguished. Based on the position of hands and elbows, we can identify occluded images.
The definition and locations of the specific keypoints used in our pipeline are shown in \cref{fig:pose_a}. Then, we can classify different situations according to the following rules:

\begin{itemize}
    \item [1)]
    \textbf{Back view}. The right shoulder ($P6$) is on the right side of the left shoulder ($P5$) on the image.
    
    \item [2)]
    \textbf{Side view}. The width to height aspect ratio $W/H$ of the person's upper body is less than 0.3. 
    
    \item [3)]
    \textbf{Occluded}\footnote{Note that only self-occlusion is considered here. Though, with the visibility probabilities predicted by HRNet we can also infer occlusion by other objects, this is not yet considered in the current pipeline.}. Any hand or elbow point ($P7$, $P8$, $P9$, $P10$) is in the upper body area (the area surrounded by $P6$, $P5$, $P11$, and $P12$) or the lower body area (the area enclosed by $P12$, $P11$, $P13$, and $P14$).
\end{itemize}

 For the width-to-height aspect ratio of the upper body (Rule 2), as shown in \cref{fig:pose_a} and \cref{fig:pose_c}, we consider the Euclidean distance between shoulders ($P5$ and $P6$) as the upper-body width $W$, and that between the center of the shoulder (the middle point of $P5$ and $P6$) and the center of the butt (the middle point of $P11$, and $P12$) as the height $H$. Then, we select qualified frontal-view images with $W/H\ge 0.3$. 
 
 For the judgment of occlusion (Rule 3), since the detected points are not on the edge of the body, we define the upper-body and lower-body areas by expanding the surrounding points. First, we define each area's width according to the top corner-point distance of that area. Specifically, as shown in \cref{fig:pose_d}, $w_1$ is the width of the upper-body area, and $w_2$ is the width of the lower-body area. Then, we extend the upper-body area and lower-body area horizontally by $w_1^{'}$=$0.1\times{w_1}$ and $w_2^{'}$= $0.1\times{w_2}$, respectively.
 
 As shown in \cref{fig:pose_result}, \cref{fig:pose_a} is a qualified frontal-view and non-occluded image. According to the position of the shoulders (Rule 1),  \cref{fig:pose_b} is a back-view image.  \cref{fig:pose_a} shows an example of $W/H\ge 0.3$, while \cref{fig:pose_c} is a side-view image because $W/H<0.3$ (Rule 2). \cref{fig:pose_d} is classified as an occluded image based on the position of hands (Rule 3).

\begin{figure*}
  \centering
  \begin{subfigure}{0.25\linewidth}
    \includegraphics[width=1\linewidth]{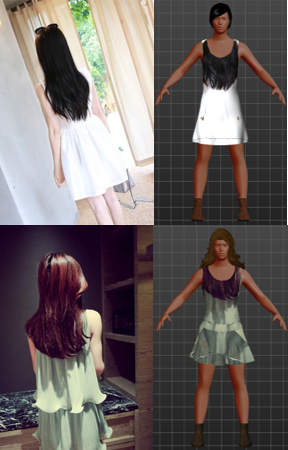}
    \caption{From back-view images}
    \label{fig:back}
  \end{subfigure}
  \hspace{5mm}
  \begin{subfigure}{0.25\linewidth}
    \includegraphics[width=1\linewidth]{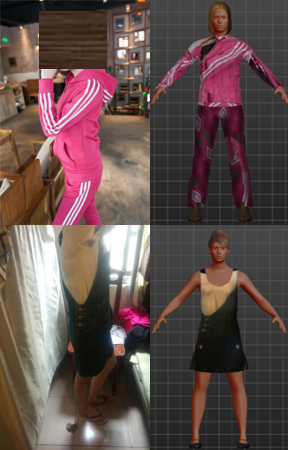}
    \caption{From side-view images}
    \label{fig:side}
  \end{subfigure}
  \hspace{5mm}
  \begin{subfigure}{0.25\linewidth}
    \includegraphics[width=1\linewidth]{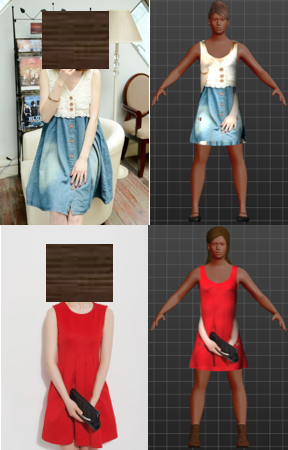}
    \caption{From occluded images}
    \label{fig:occlude}
  \end{subfigure}
  \caption{Characters created from images of different person views.}
  \label{fig:pose_result_character}
\end{figure*}

Unqualified images of person views may cause some common problems for the proposed cloning method, as shown in \cref{fig:pose_result_character}. For example, characters generated from back-view images may contain hairs (\cref{fig:back}). Characters created from side-view images may have strange textures (\cref{fig:side}). Besides, clothes occluded by hands may cause the generated characters containing ghost hands (\cref{fig:occlude}). Therefore, the proposed person view qualification step by pose detection is useful to get qualified frontal-view and non-occluded images, and thus facilitate the cloning of clean clothes.

%#############################################################################################################
\subsection{Clothes and keypoint detection}
\label{subsec:clothes_detection}

\begin{figure*}
  \centering
  \begin{subfigure}{0.8\linewidth}
    \includegraphics[width=1\linewidth]{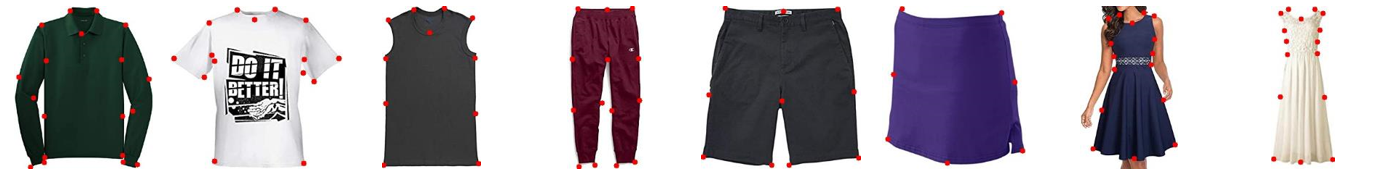}
    \caption{Eight clothes types with labeled keypoints.}
    \label{fig:texture_a_appendix}
  \end{subfigure}
  \hfill
  \begin{subfigure}{0.8\linewidth}
    \includegraphics[width=1\linewidth]{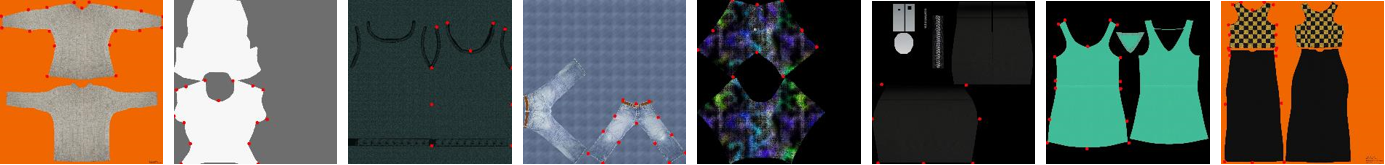}
    \caption{Regular UV maps where clothes appear in regular shapes and structures.}
    \label{fig:texture_b_appendix}
  \end{subfigure}
  \hfill
  \begin{subfigure}{0.8\linewidth}
    \includegraphics[width=1\linewidth]{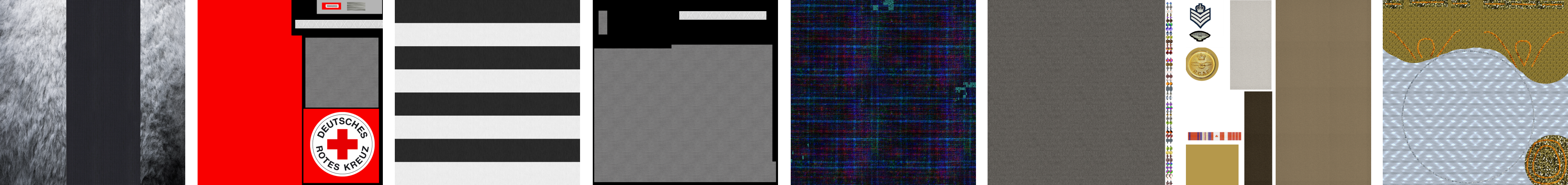}
    \caption{Irregular UV maps.}
    \label{fig:texture_c1}
  \end{subfigure}
  \caption{Different types of clothes and UV texture maps of the corresponding 3D clothes models. }
  \label{fig:textures1}
\end{figure*}

Through \cref{subsec:pedestrian} and \cref{subsec:pose}, we obtained images that contain persons' entire bodies and are completely visible. To achieve the mapping from real-world image to virtual character, we need to get the clothes position and type, and positions of the clothing keypoints in the image. Therefore, we further train two models: the clothes detection model and their corresponding key points detection model. 
% For clothes detection training, we used 191k diverse images of 13 popular clothing categories from DeepFashion2 \cite{DeepFashion2}. The clothes keypoint detection training data is composed of DeepFashion2 and crawled clothes images, in which we annotated 17k images manually. After removing the invalid images, we finally picked about 10k images of eight clothing categories that we used in this paper to train the clothes key point detection model.

\cref{fig:texture_a_appendix} shows the types of clothes we use, and the red points display positions of keypoints. The clothing models include eight models (long sleeves, short sleeves, sleeveless, trousers, shorts, skirts, short dresses, and long dresses). After obtaining the labeling information of the clothes keypoints, the clothes detection models and the keypoint detection models are trained separately for eight clothes models. The clothes detection model is based on the faster RCNN \cite{ren2015faster} which predicts the bounding box localization and clothes category jointly. The keypoint detection model is based on PIPNet \cite{JLS21}  without Neighbor Regression Module. Finally, we detect all pose qualified images and get the clothing category and the keypoint locations of clothes.

\section{Registered Clothes Mapping}

With 3D clothes models available in the MakeHuman community, we obtain some clothes models with regular UV maps, where clothes appear in regular shapes and structures, as \cref{fig:texture_b_appendix} shows. With these regular UV maps, it is possible to apply Registered Clothes Mapping to map real-world clothes textures to virtual models. However, the structure of some UV maps is not clear, so we need a way to find out its structure.

\begin{figure*}
  \centering
  \includegraphics[width=1\linewidth]{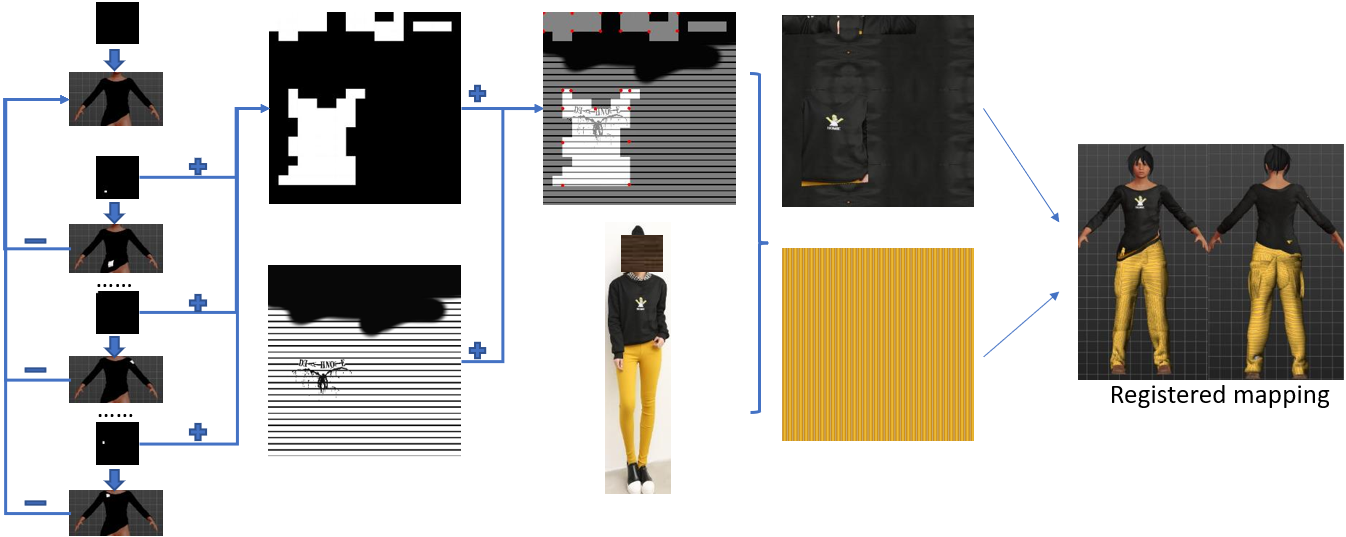}
  \caption{Find out clear structure in UV maps.}
  \label{fig:ir_to_re}
\end{figure*}

\begin{figure*}
  \centering
  \includegraphics[width=1\linewidth]{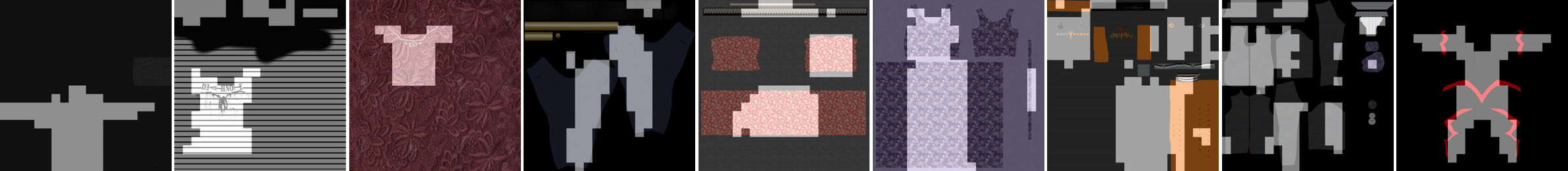}
  \caption{Examples of founded frontal areas in UV maps.}
  \label{fig:regular}
\end{figure*}

Changing the UV map will change the appearance of the 3D model because there is a correspondence between the UV map and the model. As \cref{fig:ir_to_re} shows, firstly, we use a pure black image as the UV map, and get the model's front-view image as a reference image. Next, a $50\times50$ white square is used to traverse the UV texture map and get many corresponding front-view images as response images. Then, by comparing these response images and the reference image, we could find out which area in UV maps would be mapped to the front of the model. Finally, by stacking these squares, we can get the approximate area of the texture on the front of the model. \cref{fig:ir_to_re} shows some frontal areas founded by this method. Accordingly, different region division and keypoint labeling and mapping rules are designed according to different structures of the UV maps.

% \section{Multi-view strategy}

\begin{figure}
  \centering
  \includegraphics[width=1\linewidth]{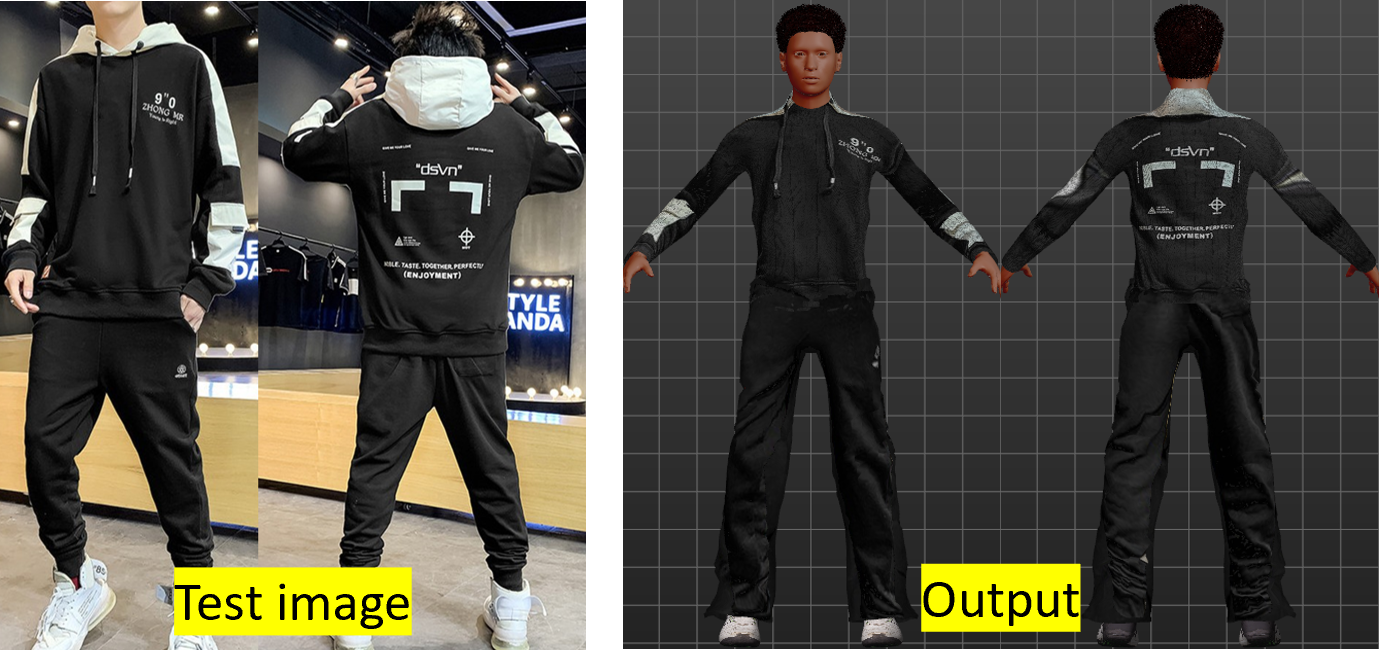}
  \caption{Character generated by multi-view images.}
  \label{fig:multi-view}
\end{figure}

\textbf{Multi-view strategy.} Note that We aim at developing a general system that requires only one single image, as multi-view images are not always available. However, when multi-view images are available as inputs, it is quite straightforward to integrate them into different parts of regular UV maps. An example is shown in \cref{fig:multi-view}.

\section{Homogeneous Cloth Expansion}

\begin{figure*}
  \centering
  \includegraphics[width=1\linewidth]{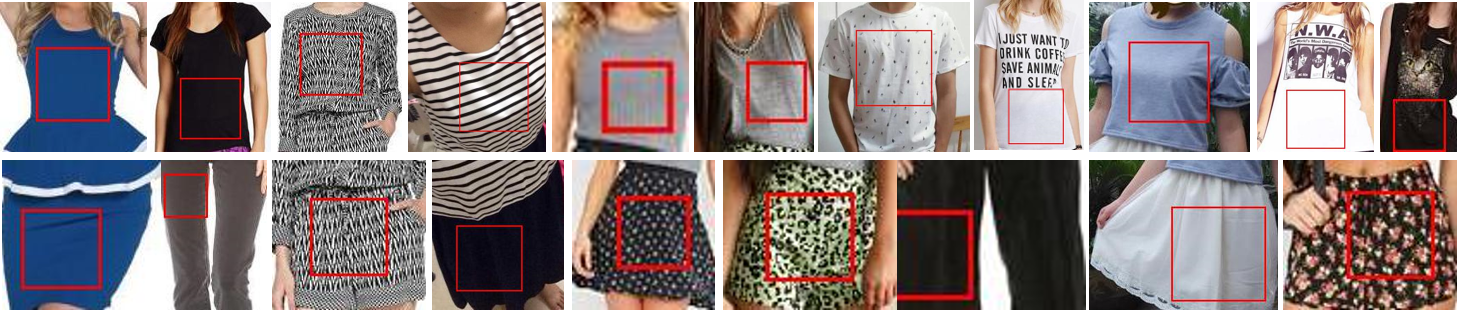}
  \caption{Examples of optimized cloth cells by the proposed algorithm.}
  \label{fig:cloth_cells}
\end{figure*}

As discussed in the main paper, to generate clothes textures for irregular UV maps, and textures on regular UV maps corresponding to invisible person parts, we further design a homogeneous cloth expansion method to find a homogeneous area on clothes as a realistic cloth cell, and expand the cell to fill the UV map. \cref{fig:cloth_cells} shows some examples of the optimized cloth cells by the proposed algorithm. From these examples, we can see that the proposed method is able to find a homogeneous cloth patch as large as possible.

\begin{figure*}
  \centering
  \includegraphics[width=1\linewidth]{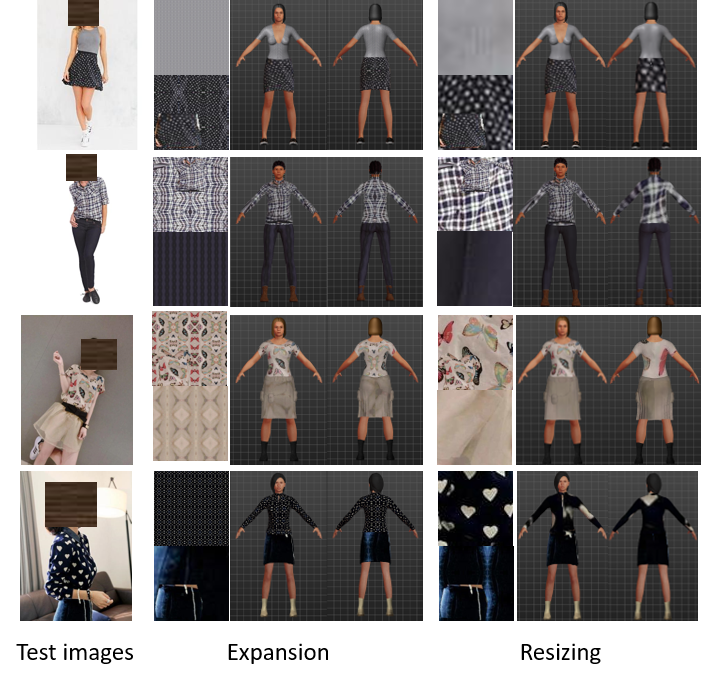}
  \caption{Comparison of expansion and resizing methods in generating UV maps and characters.}
  \label{fig:compare_expansion}
\end{figure*}

Besides the proposed homogeneous cloth expansion method, given a cropped cloth cell, a simple way to create a UV map is to resize the cloth cell directly as a UV map, as proposed in RandPerson \cite{wang2020surpassing}, and also used in UnrealPerson \cite{zhang2021unrealperson}. However, simply resizing the cloth cells may result in blur textures and unrealistic patterns. For example, \cref{fig:compare_expansion} shows a comparison between resizing and the proposed expansion methods. As can be seen, characters created by the proposed expansion method have more realistic textures, while those created by resizing are usually blur. Besides, textures created by resizing usually do not match the pattern scale of the original clothes, and thus are not able to represent the original clothes. This can also be observed from synthesized images of UnrealPerson
%, as shown in \cref{fig:unrealperson}
. In contrast, the proposed expansion method usually has a better consistency of pattern scales.

% \section{Similarity-Diversity Expansion}

% \begin{figure*}
%   \centering
%   \includegraphics[width=1\linewidth]{figs/supply6.png}
%   \caption{Example of similarity-diversity expansion.}
%   \label{fig:similarity_diversity}
% \end{figure*}

% We propose a similarity-diversity expansion strategy to scale up virtual character creation. The basic idea is illustrated in \cref{fig:similarity_diversity}. By clustering person images, we can create similar characters from the same cluster, while increase diversity by including more and more clusters. This way, the created characters can expand densely in visual similarity and diversely in population.

\section{Unity3D Simulation and Rendering}\label{sec:render}

As for the rendering process, we follow RandPerson \cite{wang2020surpassing} for the Unity3D environment settings, including the scenes, the configuration of camera networks and character movements, video capturing, and image cropping. In addition, we implement some adjustments to improve the rendering:

\textbf{Camera filter.} We set post-processing effects for some cameras to increase the imaging variations and make the data more diverse. Post-processing effects include color grading, bloom, grain, and vignette provided in Unity3D.

\textbf{Actions.} To make the generated data closer to the real-world data, we remove the running and uncommon walking actions in RandPerson. Instead, we include the situation of hanging out in place, allowing the character to stand in place and move hands or turn around, enriching the data diversity.

\begin{figure*}
  \centering
  \includegraphics[width=1\linewidth]{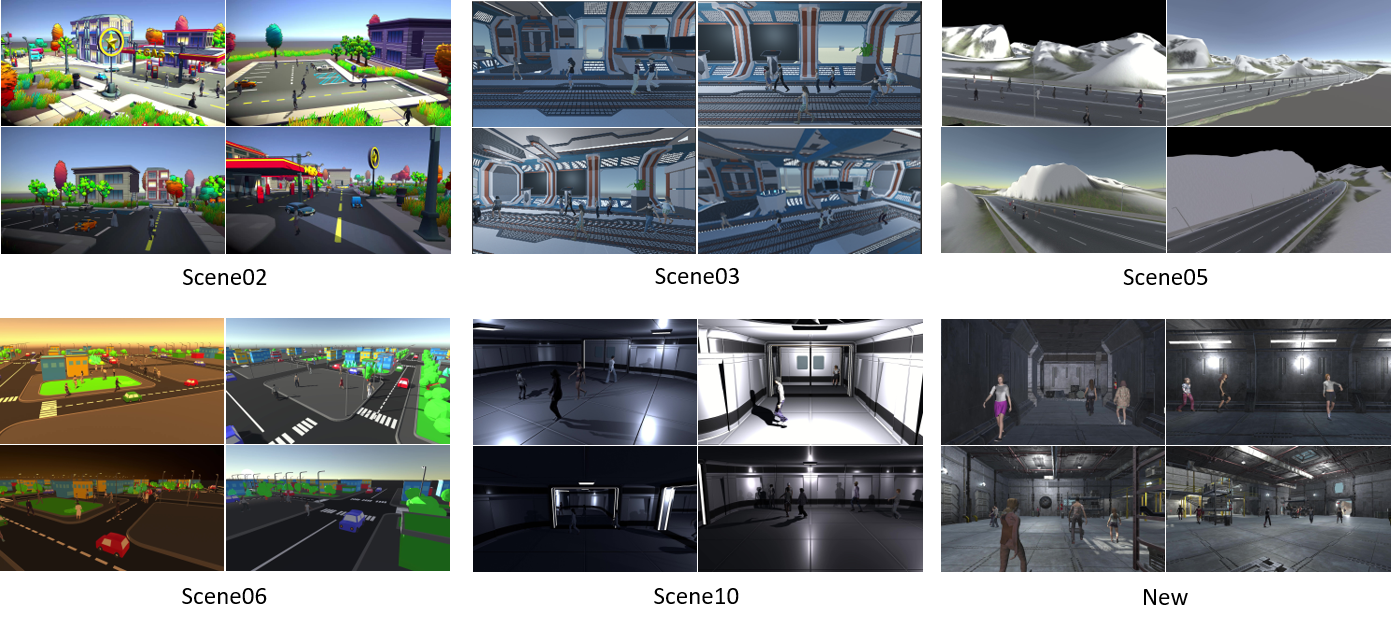}
  \caption{Unity3D virtual environments utilized in this work.}
  \label{fig:scenes}
\end{figure*}

\textbf{Scenes and cameras.} The number of cameras in each scene should be expanded to increase rendering efficiency and viewpoint diversity. Since some scenes in RandPerson are too small to expand cameras, we select five out of 11 scenes in RandPerson (scene2, 3, 5, 6, and 10) and create a new scene ourselves to get more complex lighting. We expand the number of cameras in each scene to four, making each scene's proportion in the database more balanced. In total, RandPerson uses 19 cameras in 11 scenes, while we use six scenes with 24 cameras. \cref{fig:scenes} shows the six scenes with 24 cameras we use.

\begin{figure}
  \centering
  \includegraphics[width=1\linewidth]{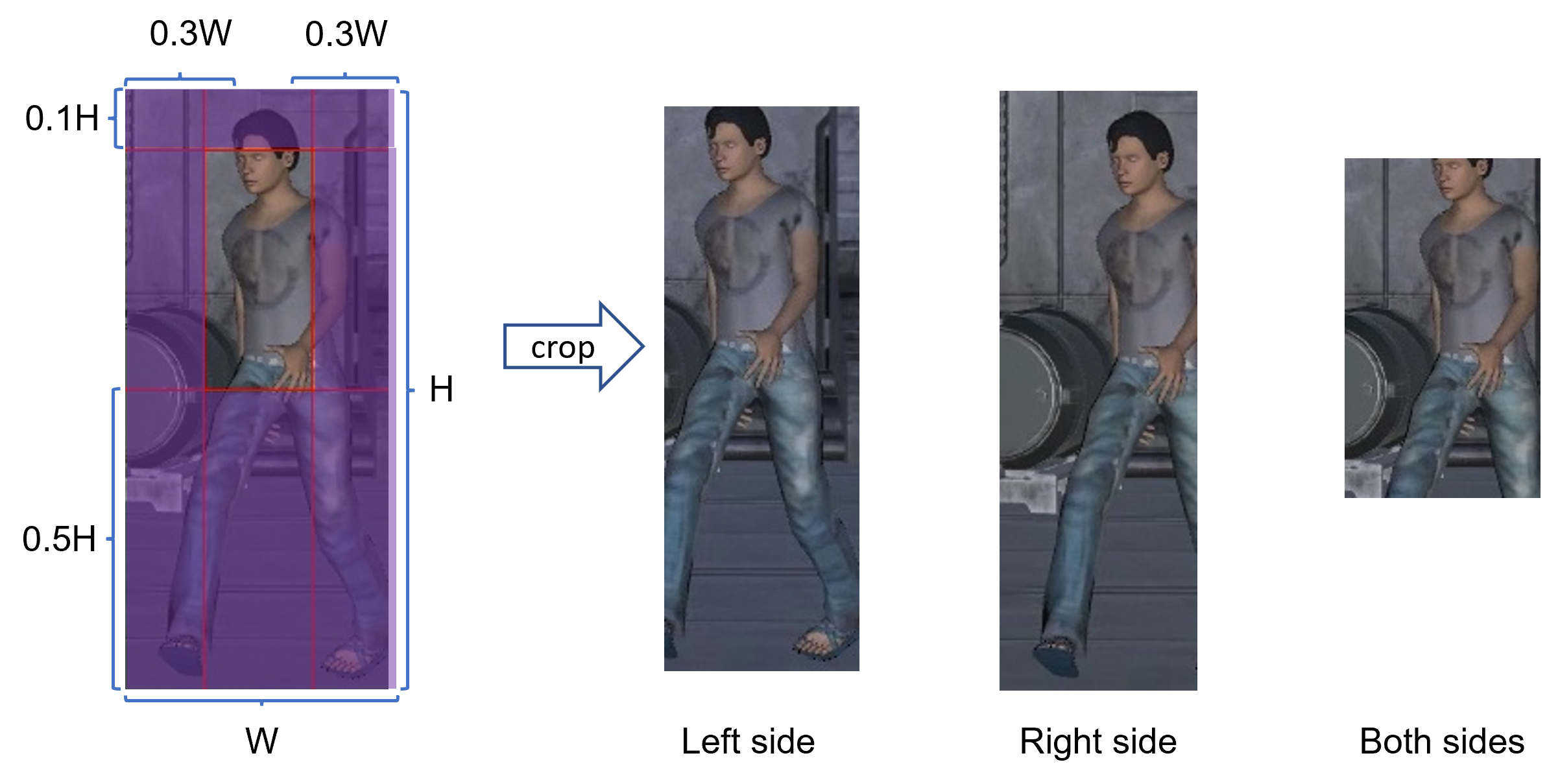}
  \caption{Illustration of image cropping. Based on the result of the RandPerson’s image cropping strategy, the occluded area on the left image shows the possible range of our random removals, and the images on the right are three examples of the cropped results.}
  \label{fig:crop}
\end{figure}

\textbf{Image cropping.} We make further improvements with RandPerson's image cropping strategy by introducing random disturbances to the cropping. Cropped persons in RandPerson are mostly complete and well-aligned. However, there are many incomplete and misaligned person images in real-world datasets. Therefore, we make random disturbances to the cropping to simulate partially visible and misaligned person images. 
Specifically, let the width and height of the original image be $W$ and $H$, respectively. For each image, with a probability $\rho$=30\% we randomly choose to further crop the image. Then, for the selected image with further cropping, we remove the top $0$-$0.1H$ part of the image randomly, and remove the bottom $0$-$0.5H$ part randomly. Then we randomly use one of the three strategies (left side only, right side only, and both sides) to remove some content randomly in $0$-$\tau W$ of the original image, where the side rate $\tau$=0.3 by default. \cref{fig:crop} illustrates the process and some cropped examples. \cref{tab:crop_result} shows the results of using different cropping strategies.

With the above setup, the generated 3D characters are imported into Unity3D environments to render and crop person images.% a virtual dataset. %Finally, we create 4,826 characters with 763,953 images, as the ClonedPerson dataset. 
% Statistics of the dataset are shown in Table \ref{tab:dataset_compare}. 
% We summarize the information of each process in the supplementary material. \cref{fig:irregular} shows some examples of the created 3D characters by the proposed approach. We also make a demonstration video of the whole process and put it in the supplementary material.

% \subsection{Scenes and Cameras.}

% We select five out of 11 scenes in RandPerson (scene2, 3, 5, 6, and 10) and create a new scene by ourselves to get more complex lighting. We expand the number of cameras in each scene to four, making each scene's proportion in the database more balanced. \cref{fig:scenes} shows the six scenes with 24 cameras we use.

% \subsection{Image cropping.}

% We make further improvements to RandPerson’s image cropping strategy by introducing random disturbances to the cropping. Specifically, let the width and height of the original image be $W$ and $H$, respectively. We randomly use one of the three strategies (left side only, right side only, and both sides) to remove some content randomly in $0$-$0.4W$ of the original image. Then we randomly select 10\% of the data, remove the top randomly in $0$-$0.1H$, and remove the bottom randomly in $0$-$0.5H$. \cref{fig:crop} illustrates the process and some cropped examples.

\section{ClonedPerson Dataset}

\begin{figure}
  \centering
  \includegraphics[width=0.9\linewidth]{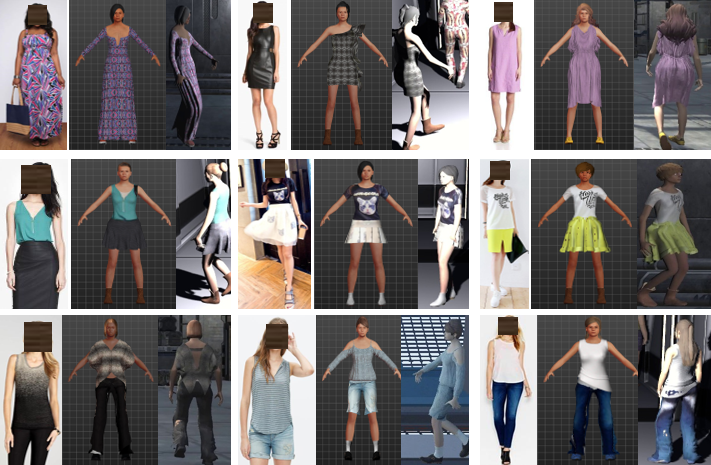}
  \caption{Examples of 3D characters in ClonedPerson. Each group contains input image, generated 3D character, and rendered person image.}
  \label{fig:irregular}
\end{figure}

An automatic pipeline is described in the main paper to directly clone the whole outfits from real-world person images to virtual 3D characters. \cref{fig:irregular} shows some examples of 3D characters in ClonedPerson. For the whole process of creating the ClonedPerson dataset, the specific information in each step is detailed as follows.
For each image, person detection needs 0.28s, pose detection needs 0.15s, clothes and keypoint detection needs 0.23s, clothes mapping and 3D creation needs 27.65s, and Unity3D rendering needs 16.5s. Therefore, for the whole pipeline each image costs 44.8s in total.

For training clothes detection, we use 191k diverse images of 13 popular clothing categories from DeepFashion2 \cite{DeepFashion2}. The clothes keypoint detection training data is composed of DeepFashion2 and crawled clothes images, in which we annotate 17k images manually. After removing the invalid images, we finally select about 10k images of eight clothing categories that we use in this paper to train the clothes detection and clothes keypoint detection models.

For cloning clothes from real-world person images to virtual characters, we use images from both DeepFashion \cite{liuLQWTcvpr16DeepFashion} and DeepFashion2, with a total of 409k images as our source data. By employing person detection, 146k person images are selected which contain detected persons. Then, 83k images are qualified by viewpoint judgment employing pose detection. Among them, 65k images are successfully detected with clothes bounding boxes and categories, as well as clothes keypoint positions. 

In the clustering stage, we use eps=0.4 to remove 29k images due to repeating persons with the same outfits. Then, we set eps=0.5, and obtain 968 clusters with 6,340 images to create characters. Among the valid 968 clusters, we use all of the clusters and select seven images in each cluster to create our 3D characters. Since some clusters are less than seven images, finally, we get 5,621 person images as inputs and create 5,621 characters accordingly by the proposed method. After rendering and cropping, we obtain 887,766 images for the 5,621 virtual persons, and this forms our ClonedPerson dataset. Among them, we use 763,953 images from 4,826 characters for training, and 123,813 images of 795 characters for testing.

\begin{figure}
  \centering
  \includegraphics[width=1\linewidth]{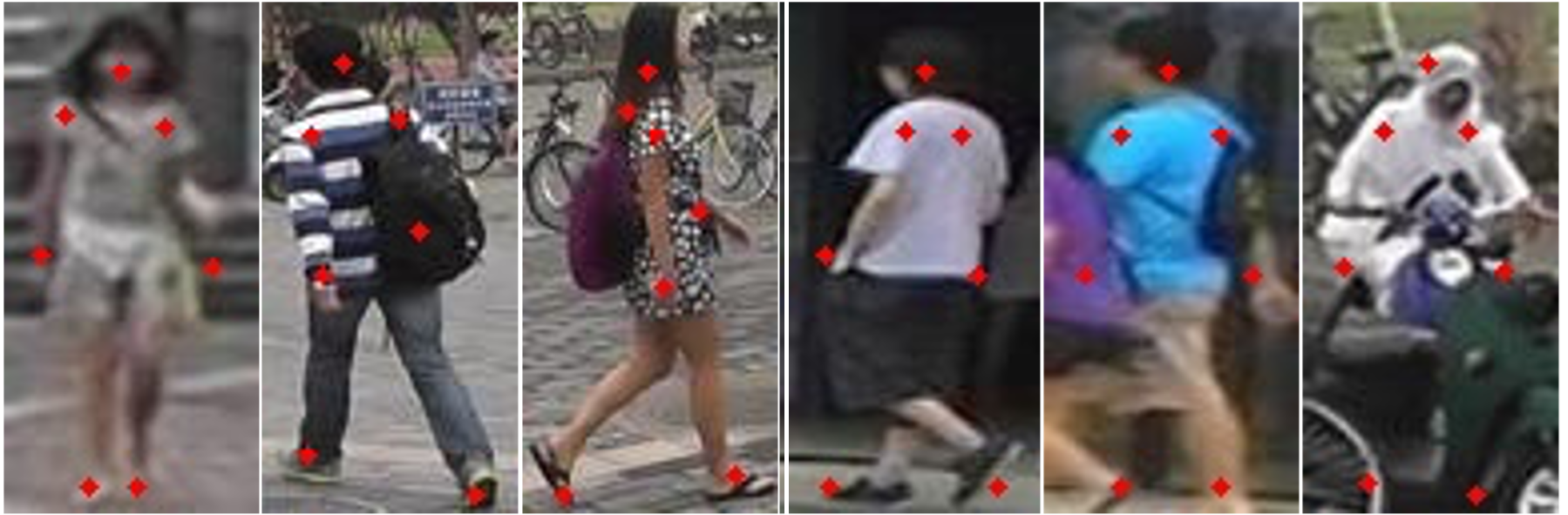}
  \caption{Illustrations of keypoint detection on real-world images with a model trained on the ClonedPerson dataset.}
  \label{fig:keypoints}
\end{figure}

Besides person re-identification, our data can also be used for other tasks e.g. person detection, person keypoint detection, multi object tracking (with videos), multi-camera multi object tracking, etc. \cref{fig:keypoints} shows some examples of person keypoint detection on real-world images with a model trained on the ClonedPerson dataset, with automatically recorded keypoint annotations.

% In the clustering stage, we use eps=0.3 to remove 2k images due to repeating persons with the same outfits. Then, we set eps=0.5 and select 5k images to create characters with fully regular UV maps for the comparison of clothes type importance, as provided in \cref{sec:compare_type}. Finally, 6k images in 820 clusters are obtained by clustering with eps=0.6, while there are still 20k images left without cluster labels. Among the valid 820 clusters, we use 80\% (656) of the total clusters and select five images in each cluster to create our 3D characters. Since some clusters are less than five images, finally, we get 3,248 images and create 3,248 characters. After rendering and cropping, we obtain 477,555 images for the 3,248 virtual persons, and this forms our ClonedPerson dataset.

% \section{3D Virtual Character Generation}

\section{EXPERIMENTS}

\subsection{Comparison of Different Cropping Strategies}
\begin{table}[b] 
 \begin{center}
  \begin{tabular}{@{}c@{}|@{}c@{}|c|c|c|c}
   \hline
        Prob. $\rho$ & Side Rate $\tau$ & \#ID & \#Images & Rank-1 & mAP \\
   \hline\hline
        0&0&	4,826&	    763,953&	45.7&	26.7\\
        10\%&0&	4,826&	    763,953&	48.9&	29.9\\
        20\%&0&	4,826&	    763,953&	48.7&	29.8\\
        30\%&0&	4,826&	    763,953&	49.0&	30.1\\
        40\%&0&	4,826&	    763,953&	48.8&	30.0\\
        50\%&0&	4,826&	    763,953&	48.8&	29.9\\
        30\%&10\%&	4,826&	    763,953&	49.7&	31.1\\
        30\%&20\%&	4,826&	    763,953&	51.2&	32.3\\
        30\%&30\%&	4,826&	    763,953&	52.1&	33.4\\
        30\%&40\%&	4,826&	    763,953&	51.3&	32.7\\
   \hline
  \end{tabular}
   \caption{Results of different cropping strategies with the cropping probability $\rho$ and side rate $\tau$.}
   \label{tab:crop_result}
 \end{center}
\end{table}

\cref{tab:crop_result} shows the performance of different cropping strategies with the cropping probability $\rho$ and side rate $\tau$ as introduced in \cref{sec:render}. Firstly, we only change the cropping probability $\rho$. From the results shown in \cref{tab:crop_result}, it can be observed that the best result is achieved with $\rho$=30\%. Then, we keep $\rho$=30\%, and change the side rate $\tau$. Finally, from \cref{tab:crop_result} it can be observed that it achieves the best performance with the cropping probability $\rho$=30\% and the side rate $\tau$=30\%. Therefore, the two values are kept as default values.

\subsection{Comparison to Existing Datasets}

Due to space limits of the main paper, we report the detailed results of different datasets for different tasks in  \cref{tab:detail_tasks}.
\begin{table*}

 \begin{center}
  \begin{tabular}{c|c|c|c|c|c|c|c|c|c|c|c}
   \hline
     \multirow{2}{*}{Method}&\multirow{2}{*}{Dataset}&\multirow{2}{*}{\#ID} & \multirow{2}{*}{\#Imgs}& \multicolumn{2}{c|}{CUHK03-NP} & \multicolumn{2}{c|}{Market-1501} &  \multicolumn{2}{c|}{MSMT17}&  \multicolumn{2}{c}{Average}  \\
    \cline{5-12}
        &&&& Rank-1 & mAP & Rank-1 & mAP & Rank-1 & mAP& Rank-1 & mAP \\
   \hline\hline
   \multirow{6}*{QAConv}& RandPerson& 8,000 & 1,801k&17.8& 16.0 &74.5& 46.9 &40.6& 14.0&44.3&25.6\\
    ~&RandPerson &8,000& 132k & 16.8&15.1&75.9&45.9&40.8&13.8&44.5&24.9\\
        ~&RandPerson$^*$ & 8,000& 1,239k & 20.5&20.1&81.6  &56.4&46.8   &17.6&49.6&31.4\\
    ~ &UnrealPerson & 6,799&1,256k  &19.2& 17.2 &80.0& 56.1 &46.0& 17.5&48.4&30.3\\
     ~ &UnrealPerson &3,000&120k&18.8&17.8&80.6  &55.9&\textbf{49.5}  &\textbf{19.3}&49.6&31.0\\

    ~ &ClonedPerson & 4,826 & 763k &\textbf{22.6}&\textbf{21.8} &\textbf{84.5}& \textbf{59.9}&49.1&18.5&\textbf{52.1}&\textbf{33.4}\\
    \hline\hline
      \multirow{8}*{TransMatcher} &Market\cite{liao2021transmatcher}&- & -& 22.2&21.4&- & -&47.3&18.4&-&-\\ 
      ~ &MSMT17\cite{liao2021transmatcher}&- & -& 23.7&22.5&80.1 & 52.0&-&-&-&-\\ 
      ~ &RandPerson&8,000 & 1,801k& 21.2&18.7&77.6 & 49.6&45.3&16.4&48.0&28.2\\ 
      ~& RandPerson&8,000 & 132k&18.4&16.9&77.3& 49.0&44.3&15.8&46.7&27.2\\
       ~&RandPerson$^*$ & 8,000& 1,239k & 22.9&22.9&83.6  &58.0&51.4   &20.9&52.6&33.9\\
        ~ &UnrealPerson & 6,799&1,256k  &21.8& 19.7 &81.1& 60.2&44.8&18.4 &49.2&32.8\\
      ~ & UnrealPerson & 3,000&120k  & 21.4&19.6 &81.6& 59.4&\textbf{52.0}&\textbf{21.6} &51.7&33.5\\
  ~ & ClonedPerson& 4,826 & 763k &\textbf{25.4}& \textbf{24.4} &\textbf{84.8}& \textbf{62.3}&51.6&20.8&\textbf{53.9}&\textbf{35.8}\\
  
     \hline\hline
%   Msmt(SpCL) & 1,041 &30,248&  && \\
  \multirow{3}*{SpCL} &RandPerson& 8,000 & 132k&3.9&4.7  &83.4&67.2&53.7&27.2&47.0&33.0\\
    ~ &UnrealPerson & 3,000&120k  &4.2&5.3 &86.5&71.7 &\textbf{55.2}&\textbf{28.4}&48.6&35.1\\
  ~ &ClonedPerson & 4,826 & 75k &\textbf{11.7}&\textbf{ 12.0 }&\textbf{88.0}&\textbf{72.7}&49.3&24.2&\textbf{49.7}&\textbf{36.3}\\
   \hline
  \end{tabular}
   \caption{Results with different datasets for different tasks. RandPerson$^*$ means an adapted RandPerson dataset rendered with the same settings of the ClonedPerson.}
   \label{tab:detail_tasks}
 \end{center}
\end{table*}

\section{Limitations}

\begin{figure*}
  \centering
  \includegraphics[width=0.9\linewidth]{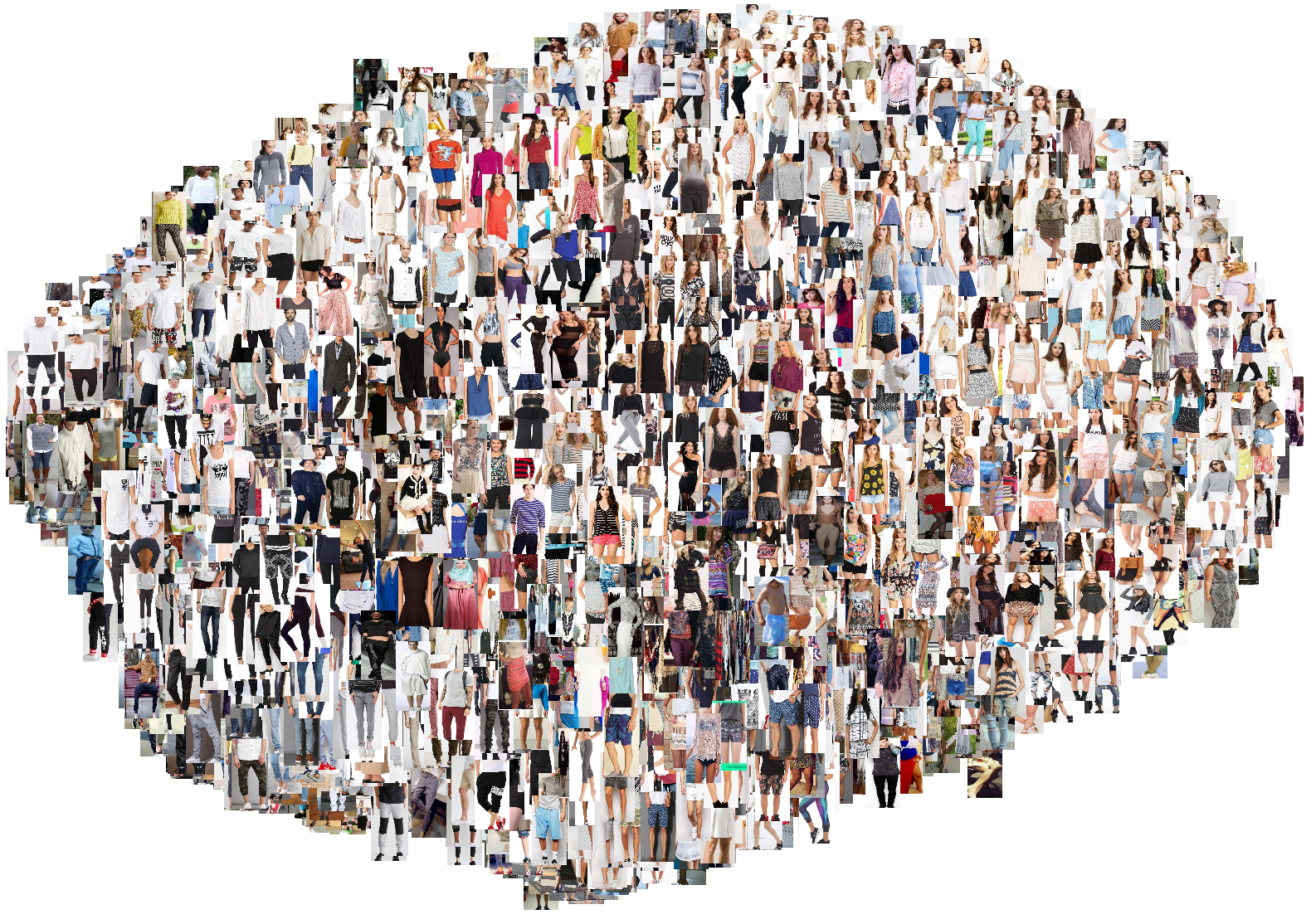}
  \caption{Distribution of DeepFashion and DeepFashion2 images, made by t-SNE \cite{van2008visualizing}.}
  \label{fig:deepfashion_analysis}
\end{figure*}

As summarized below, this research leaves some aspects for improvements. 

\begin{itemize}
    \item[(1)] Limited virtual character clothes models. The models we used are from the MakeHuman community, where the available models are limited. Because of the limited models, the categories of clothes can be applied are thus limited. 
    % Although we further use 186 models with irregular UV maps to compensate this shortcoming as much as possible, by irregular models we cannot clone global textures of clothes well. The results shown in \cref{tab:clothes_types} also indicate that limited number of regular models suppresses its performance. Therefore, in the future, we need to exploit more clothes models with regular UV maps.
    
    \item[(2)] Limited data source. We mainly use images from DeepFashion and DeepFashion2 datasets to create our virtual characters. This makes the data source not diversified enough. We show a distribution of the DeepFashion and DeepFashion2 images in \cref{fig:deepfashion_analysis}. We use the same model trained on MSMT17 by QAConv 2.0 to compute similarity scores between images, and draw a sample distribution by t-SNE \cite{van2008visualizing}. By this plot, we can find that clothes in DeepFashion datasets are not diversified enough. For example, most of the images are summer clothes in white or black. Therefore, we need to exploit more data sources in the future.
    
    \item[(3)] Only clothes considered. The proposed method only clones clothes from person images, but is not capable of high-fidelity reconstruction of 3D models from person images. However, our motivation is to create diversified characters with realistic clothing and create a dataset for improved generalization. High-fidelity reconstruction is challenging and not efficient for our purpose. On the other hand, high-fidelity reconstruction of identifiable biometric signatures, e.g. faces, may also raise privacy concerns.
\end{itemize}

%%%%%%%%% REFERENCES
{\small
\bibliographystyle{ieee_fullname}
\bibliography{appendix}
}
% \end{document}
% \end{appendices}